\documentclass[11pt]{article}

\usepackage[preprint]{acl}
\usepackage{times}
\usepackage{latexsym}

\usepackage[T1]{fontenc}
\usepackage[utf8]{inputenc}
\usepackage{microtype}
\usepackage{inconsolata}
\usepackage{graphicx}
\usepackage{algorithm}
\usepackage{algpseudocode}
\usepackage{amssymb}
\usepackage{amsmath}
\usepackage{booktabs}
\usepackage{url}
\usepackage{multirow}
\usepackage{array}
\usepackage[table]{xcolor}
\usepackage{makecell}
\usepackage{cleveref}
\usepackage{unicode-math}
\usepackage{tcolorbox}
\tcbuselibrary{breakable}
\makeatletter
\newenvironment{compactalgorithm}[1][t]{%
  \let\origfsruled\fs@ruled
  \def\fs@ruled{%
    \origfsruled
    \def\@fs@post{%
      \kern2pt\hrule\relax\vskip-1em\relax
    }%
  }%
  \begin{algorithm}[#1]%
}{%
  \end{algorithm}%
}
\makeatother

\newcommand{\method}{HERO}

\title{\method: Harnessing Evidence for Test-time Refinement and Prompt Optimization in Program Generation}

\author{
  \textbf{Mingyu Lu}\textsuperscript{1,2}\thanks{Work done during internship at Microsoft Research.} \quad
  \textbf{Timothy Ossowski}\textsuperscript{2} \quad
  \textbf{Tristan Naumann}\textsuperscript{2} \\
  \textbf{Hoifung Poon}\textsuperscript{3} \quad
  \textbf{Peniel Argaw}\textsuperscript{2} \\
  \\
  \textsuperscript{1}University of Washington \quad
  \textsuperscript{2}Microsoft Research \quad
  \textsuperscript{3}Recursion \\
  \texttt{mingyulu@cs.washington.edu} \\
}

\begin{document}
\maketitle
\begin{abstract}
Harness optimization improves LLMs by refining reusable prompts and scaffolding, yet requires further adaptation for program generation tasks such as text-to-SQL and text-to-Python.
Existing methods often rely on LLM reflectors to propose harness updates, which can fix observed failures but also disrupt previously correct behavior. Text-to-code tasks offer a natural way to mitigate this issue: we can execute generated outputs, allowing us to evaluate and verify the effects of proposed updates directly. Building on this signal, we introduce \textsc{HERO} (\textbf{H}arnessing \textbf{E}vidence for \textbf{R}efinement and \textbf{O}ptimization), which combines offline and test-time harness optimization. At the first stage, \method{} learns reusable prompt updates from execution-verified successes and failures and validates candidate revisions before adoption. At test time, we treat the generated code as an instance-specific harness and refine it using execution feedback. Across four SQL, Cypher, and Python benchmarks, \method{} achieves the best offline performance, improving pass@1 by \(7.6\) points over the initial prompt, while test-time refinement adds another \(2.0\) points and improves execution efficiency.


\end{abstract}

\section{Introduction}
LLM agents are increasingly employed to interact with external environments through tools, often by generating executable programs such as SQL and Cypher queries and Python functions \citep{li2023can,schick2023toolformer,feng2025cypherbench,zhuo2025bigcodebench,ouyang2026dscodebench}. These artifacts provide a flexible interface for models to perform computation and interact with external systems \citep{gao2023pal,wang2024executable}. Their quality depends not only on the underlying model, but also on the surrounding \emph{harness}---the prompts, tools, and memory that shape generation \citep{khattab2024dspy,lee2026meta,li2026agent,lopopoloharness}. This makes the harness a natural target for improving LLM agent performance. Accordingly, existing work has pursued two related directions: optimizing reusable harness components across tasks \citep{lee2026meta,lin2026agentic,yang2026skillopt,agrawal2026gepa}, and using test-time feedback to select or refine generated outputs \citep{ni2023lever,chen2024teaching,madaan2023self}.


Improving program generation through these directions presents two linked challenges. First, reusable updates must correct recurring errors without degrading behavior that already works. LLM-generated feedback can be unreliable without external grounding \citep{huang2024large,kamoi2024can}, and prompt revisions that resolve failures can introduce regressions on previously successful cases \citep{wu2024strago}. The challenge is therefore to derive grounded diagnoses of what to change and preserve from execution outcomes, and translate these diagnoses into reusable guidance. Second, this general guidance must be complemented by feedback specific to the current instance. An optimized prompt captures strategies learned across training tasks, but a generated program may still encounter errors that require observations from its own execution. Approaches such as Dynamic Cheatsheet (DC) and ACE accumulate experience through evolving shared contexts \citep{suzgun2026dynamic,zhang2026agentic}. When shared context is updated at test time, unreliable or overly specific feedback can become persistent guidance, allowing errors from one instance to affect subsequent tasks. Together, these challenges motivate grounding offline harness updates in execution evidence while keeping test-time corrections local to the current program.



To address these challenges, we introduce \textsc{HERO} (\textbf{H}arnessing \textbf{E}vidence for \textbf{R}efinement and \textbf{O}ptimization), a two-level framework that grounds reusable guidance updates in execution evidence and confines test-time adaptation to the current program. Our contributions are threefold:
\begin{itemize}
    \item \textbf{A unified two-level formulation.}
    We formulate program generation as a two-level optimization problem that couples offline prompt optimization with test-time refinement of instance-specific programs.
    \item \textbf{Execution-grounded feedback and verification.}
    Offline, we use execution evidence from successful and failed programs under prior prompts to identify what to correct and preserve. At test time, execution feedback and retrieved verified training examples guide program revisions, which are accepted only when supported by evidence.
    \item \textbf{Empirical evaluation and analysis.}
    We demonstrate correctness gains over existing offline harness optimizers across diverse query-generation and programming tasks, with further improvements from test-time refinement. We further evaluate execution efficiency and computational cost, finding that the learned prompts remain compact across tasks.
\end{itemize}


\begin{figure*}
    \centering
    \includegraphics[width=0.9\linewidth]{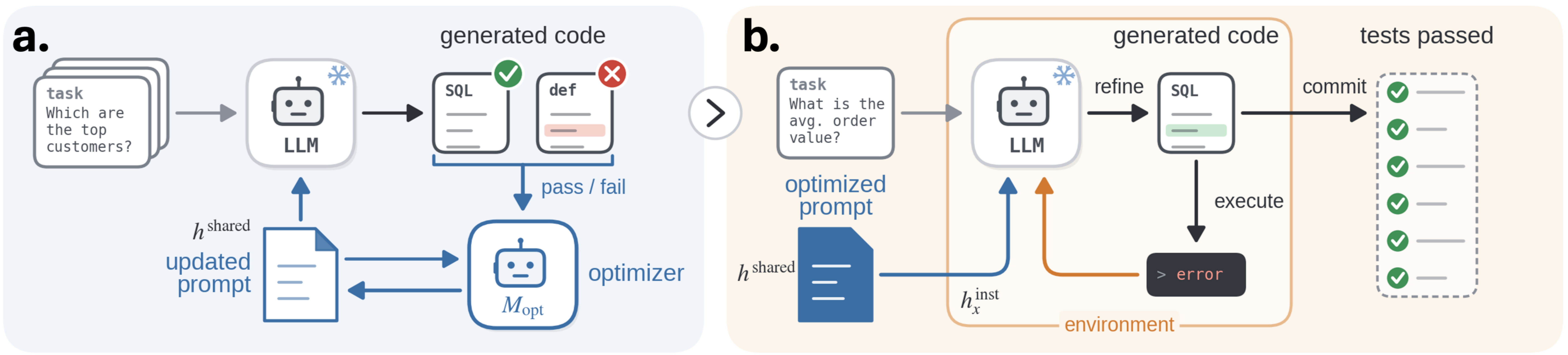}
    \caption{\textbf{Overview of \method{}.}
    \textbf{(a)} \method{}-O: Offline prompt optimization learns a shared prompt across training tasks. \textbf{(b)} \method{}-T: Test-time refinement refines the generated artifact using environment feedback.}    
    \label{fig:concept}
    \vspace{-1em}
\end{figure*}

\vspace{-0.5em}

\section{Related Works}
\noindent \textbf{Prompt and harness optimization.}
Prompt and harness optimization improves frozen LLMs by adapting reusable components around them. MIPRO \citep{opsahl2024optimizing} jointly optimizes instructions and few-shot demonstrations using task-level feedback. Reflection-based approaches include GEPA \citep{agrawal2026gepa}, which uses trajectory feedback to propose prompt revisions, and STRAGO \citep{wu2024strago}, which derives refinement strategies from both successful and failed examples to mitigate prompt drift. Other approaches broaden the optimization target: ACE \citep{zhang2026agentic} maintains an evolving context through generation, reflection, and curation; SkillOpt \citep{yang2026skillopt} optimizes reusable textual skills; and Meta-Harness \citep{lee2026meta} searches over harness implementations using source code, scores, and execution traces. Building on this line of work, \method{} adopts a similar offline optimization loop but enriches reflection with execution feedback from successful and failed programs generated under prior prompt revisions, aiming to improve program generation while limiting regressions.

\noindent \textbf{Test-time refinement.}
Test-time methods refine model outputs using feedback available during inference. ReAct \citep{yao2022react} interleaves reasoning, actions, and environment observations, while Reflexion and related self-refinement methods use feedback from earlier attempts to revise subsequent outputs \citep{shinn2023reflexion,madaan2023self}. Dynamic Cheatsheet (DC) \citep{suzgun2026dynamic} further accumulates self-curated strategies across queries through persistent memory. These methods emphasize adaptation from test-time experience. \method{} complements this adaptation with reusable guidance learned through offline prompt optimization. The optimized prompt remains fixed during inference, while execution feedback and retrieved, verified training examples guide refinement of the current program.

\vspace{-0.5em}
\section{Harness Optimization for Program Generation}
\label{sec:problem_setup}
\subsection{Problem Setup: Program Generation}
In this work, we focus on program generation---tasks that map natural language specifications to programs, including text-to-SQL, text-to-Cypher, and text-to-Python function generation. Each instance \(x\) contains a specification and available task context, such as a database schema or function signature. An LLM generates a candidate program \(y\), whose execution yields observations such as runtime errors, returned values, and execution traces \citep{wang2024executable}. We denote its task performance by an execution-based evaluation reward \(R(y,x)\in\mathbb{R}\). We distinguish this evaluation reward from the feedback available during refinement: in our setting, test-time refinement can access execution observations but not test reference solutions or hidden correctness labels.

\subsection{A Unified View of Harness Optimization}
Prior work improves fixed LLMs through reusable prompt or harness optimization \citep{agrawal2026gepa,lee2026meta} and feedback-driven refinement across attempts \citep{shinn2023reflexion}. We represent these forms of adaptation through a harness comprising external state that conditions a fixed policy \(\pi_\theta\) or mediates its interaction with the environment. For an instance \(x\), we decompose this state as
\begin{equation}
    h_x = \left(h^{\mathrm{shared}}, h_x^{\mathrm{inst}}\right),
    \label{eq:joint_harness}
\end{equation}
where \(h^{\mathrm{shared}}\) is reused across instances and \(h_x^{\mathrm{inst}}\) contains instance-specific state. During refinement, previously generated programs, their execution feedback, and retrieved examples can become part of \(h_x^{\mathrm{inst}}\), conditioning subsequent revisions. Given the harness, the policy generates the next candidate program:
\begin{equation}
    y \sim \pi_\theta(\cdot \mid x; h_x).
\end{equation}
Optimizing shared state targets the expected reward across instances, whereas optimizing instance-specific state targets the current instance.

\vspace{-0.5em}

\section{Method}\label{sec:method}
\vspace{-0.5em}

Building on this unified view, we propose \method{}, a two-stage framework that optimizes the harness at complementary scopes while keeping the underlying LLM fixed. The first stage, offline optimization (HERO-O), learns a shared harness \(h^{\mathrm{shared}}\) from execution feedback across training tasks (\Cref{fig:concept}(a)). The second stage, test-time refinement (HERO-T), keeps the shared harness fixed while adapting an instance-specific harness \(h_x^{\mathrm{inst}}\) using evidence from the current test instance (\Cref{fig:concept}(b)). The overall procedure is summarized in \Cref{alg:harness-optimization}.

\vspace{-0.5em}
\subsection{HERO-O: Offline Prompt Optimization}
\label{sec:stage1}
\newcommand{\algstage}[2]{%
    \Statex
    \hspace{-\algorithmicindent}%
    \begingroup
    \setlength{\fboxsep}{2pt}%
    \colorbox{black!5}{%
        \parbox{\dimexpr\linewidth+\algorithmicindent-2\fboxsep\relax}{%
            \strut\textbf{#1: #2}%
        }%
    }%
    \endgroup
}

\begin{compactalgorithm}[t]
\small
\caption{\method{}}
\label{alg:harness-optimization}
\begin{algorithmic}[1]
\Require Initial prompt $h_0^{\mathrm{shared}}$;
diagnostic set $\mathcal{D}_{\mathrm{diag}}$;
update set $\mathcal{D}_{\mathrm{upd}}$;
selection set $\mathcal{D}_{\mathrm{sel}}$;
optimizer LLM $M_{\mathrm{opt}}$;
candidates per iteration $N$;
iterations $T$;
program-refinement operator $\Phi_B$ with per-query action budget $B$

\algstage{Stage 1}{Prompt Optimization}

\State $\mathcal{A} \gets \{h_0^{\text{shared}}\}$;\quad
       $\mathcal{R}_{-1} \gets \emptyset$

\For{$t=0,\ldots,T-1$}
    \State $F_t \gets
    {M_{\mathrm{opt}}}
    (h_t^{\text{shared}},\mathcal{R}_{t-1},\mathcal{D}_{\mathrm{diag}})$
    \Comment{reflect on diagnostic programs and execution outcomes}

    \State $\mathcal{H}_t \gets
    \{M_{\mathrm{edit}}^{(i)}(h_t^{\text{shared}},F_t)\}_{i=1}^{N}$
    \Comment{propose candidate prompts}

    \State $(h_{t+1}^{\text{shared}},\mathcal{H}_t^{+}) \gets \Call{ScreenAndAccept} {h_t^{\text{shared}},\mathcal{H}_t,\mathcal{D}_{\mathrm{upd}}}$
    \Comment{\Cref{alg:candidate-acceptance}}

    \State $\mathcal{R}_t
    \gets \mathcal{H}_t \setminus \{h_{t+1}^{\text{shared}}\}$

    \State $\mathcal{A}
    \gets \mathcal{A}\cup\mathcal{H}_t^{+}$
\EndFor

\State $\hat h^{\mathrm{shared}} \gets
\arg\max_{h\in\mathcal A}
\operatorname{Acc}(h,\mathcal D_{\mathrm{sel}})$


\algstage{Stage 2}{Test-Time Refinement}
\ForAll{test instances $x$}
    \State $y_0(x) \sim
    \pi_\theta(\cdot \mid x;\hat h^{\mathrm{shared}})$
    \State $\hat y(x) \gets
    \Phi_B(x,y_0(x);\hat h^{\mathrm{shared}})$
\EndFor

\State \Return $\hat h^{\text{shared}}$ and refined programs $\{\hat y(x)\}$
\end{algorithmic}
\end{compactalgorithm}

Stage~1 of \method{} instantiates the \(h^{\mathrm{shared}}\) optimization by learning a reusable prompt across training tasks. We denote \(h_t^{\text{shared}}\) for the accepted prompt at iteration \(t\). Starting from \(h_0^{\text{shared}}\), optimization alternates between diagnosing the current prompt, proposing revisions, and validating them on held-out examples. To prevent overfitting, we use three data splits with distinct roles: \(\mathcal{D}_{\mathrm{diag}}\) provides evidence for proposing revisions, \(\mathcal{D}_{\mathrm{upd}}\) determines whether a revision is accepted, and \(\mathcal{D}_{\mathrm{sel}}\) selects the final prompt. Thus, examples used to propose an update are not used to determine its acceptance.

\makeatletter
\newcommand{\algorithmbottomnote}[2][1em]{%
    \let\alg@ruled\fs@ruled
    \def\fs@ruled{%
        \alg@ruled
        \let\alg@post\@fs@post
        \def\@fs@post{%
            \alg@post
            \kern3pt
            \hbox{\parbox{\linewidth}{\scriptsize #2}}%
            \vskip-#1\relax
        }%
    }%
}
\makeatother

\begingroup
\algorithmbottomnote{%
\(\dagger\): In this algorithm, $h_t$ denotes $h_t^{\mathrm{shared}}$.%
}

\begin{algorithm}[t]
\small
\caption{\small \textproc{ScreenAndAccept} - line 5 of Algorithm 1}
\label{alg:candidate-acceptance}
\begin{algorithmic}[1]
\Require Current prompt $h_t^{\dagger}$; candidates $\mathcal{H}_t$;
update set $\mathcal{D}_{\mathrm{upd}}$ partitioned into chunks
$C_1,\dots,C_K$; budget $b$;
margin $\tau$; confidence multiplier $z$

\State Pull $C_1$ for all $h\in\mathcal{H}_t$
\While{$b>0$ \textbf{and} not \Call{Resolved}{$\mathcal{H}_t$}}
    \State $(\hat\Delta_h,\hat\sigma_h)\gets
    \Call{PairedDelta}{h,h_t}$ for all $h\in\mathcal{H}_t$
    \State $\ell\gets\arg\max_h\hat\Delta_h$;\quad
    $c\gets\arg\max_{h\neq\ell}(\hat\Delta_h+z\hat\sigma_h)$
    \State $\Call{Pull}{\ell,C_{n_\ell+1}}$;\quad $b\gets b-1$
    \If{$b>0$ \textbf{and} $\hat\Delta_c+z\hat\sigma_c>\tau$}
        \State $\Call{Pull}{c,C_{n_c+1}}$;\quad $b\gets b-1$
    \EndIf
\EndWhile
\State $(\hat\Delta_h,\hat\sigma_h)\gets
\Call{PairedDelta}{h,h_t}$ for all $h\in\mathcal{H}_t$
\State $\ell\gets\arg\max_h\hat\Delta_h$
\State $h_{t+1}\gets
\begin{cases}
\ell, & \hat\Delta_\ell>\tau,\\
h_t, & \text{otherwise}
\end{cases}$
\State $\mathcal{H}_t^+\gets
\{h\in\mathcal{H}_t:\hat\Delta_h>\tau\}$
\State \Return $h_{t+1},\mathcal{H}_t^+$
\end{algorithmic}
\end{algorithm}

\subsubsection{Textual Gradients from Execution-grounded Experience}
Prompt optimization can be viewed as a discrete analogue of gradient-based optimization, where natural language critiques or reflections serve as ``textual gradients'' for prompt updates
\citep{pryzant2023automatic,yuksekgonul2024textgrad}. 
Prior work uses LLM reflection to guide prompt updates
\citep{pryzant2023automatic,lin2026agentic,agrawal2026gepa},
but such feedback can be unreliable without external grounding
\citep{huang2024large,kamoi2024can}. Following \citet{wu2024strago}, we use both successful and failed cases but enrich reflection with execution outcomes and traces from programs generated under prior prompt revisions to identify what to correct and preserve. This connects prior edits to observed program behavior, providing concrete context for diagnosing what to correct and preserve. At iteration \(t\), the optimizer LLM reviews the prompt edits proposed at \(t-1\), along with \(k\) generated programs from both successful and failed cases and their corresponding execution outcomes (e.g., failure errors or predicted output mismatches). Correct programs indicate behaviors to preserve, while incorrect programs identify behaviors to revise. The optimizer summarizes this evidence into feedback \(F_t\), providing an execution-grounded direction for updating the current \(h_t^{\text{shared}}\).

Conditioned on \(h_t^{\text{shared}}\) and \(F_t\), the same optimizer LLM proposes \(N\) candidate shared harnesses,
\begin{equation}
    \mathcal{H}_t
    =
    \left\{
        M_{\mathrm{opt}}^{(i)}(h_t^{\text{shared}},F_t)
    \right\}_{i=1}^{N}.
\end{equation}
We instruct the optimizer to propose revisions as reusable rules rather than instance-specific fixes, and reject candidates that copy instance-specific content or reference answers using deterministic filters, e.g., filtering out prompts that memorize training answers.

\subsubsection{Adaptive Paired Acceptance}
Prior work has explored bandit-based candidate selection \citep{pryzant2023automatic} and minibatch comparisons of prompt revisions against their parent prompts \citep{agrawal2026gepa}. Our acceptance procedure combines these strategies to improve the reliability and efficiency of prompt updates: paired comparisons capture both fixes and regressions relative to the current prompt, uncertainty-guided allocation focuses evaluation on uncertain candidates, and a minimum improvement margin guards against accepting marginal gains (\Cref{alg:candidate-acceptance}). For each candidate prompt \(h \in\mathcal{H}_t\), we compute the per-instance improvement
\begin{equation}
    d_h(x)
    =
    R(y_h,x)-R(y_{t},x),
    \label{eq:paired-delta}
\end{equation}
where $y_h\sim\pi_\theta(\cdot\mid x;h)$ and
$y_t\sim\pi_\theta(\cdot\mid x;h_t^{\mathrm{shared}})$
are generated by the candidate and current shared prompts,
respectively, on the same instance $x$. For the binary correctness reward used during acceptance, \(d_h(x)\in\{-1,0,+1\}\), corresponding to a regression, unchanged correctness, or a fix. The mean paired improvement therefore measures net fixes per evaluated instance. We denote its estimate by \(\hat{\Delta}_h\) and its estimated standard error by \(\hat{\sigma}_h\).

All candidates are initially evaluated on a common subset of \(\mathcal{D}_{\mathrm{upd}}\). At each allocation step, we identify the leader with the largest \(\hat{\Delta}_h\) and the alternative with the highest upper confidence bound (UCB)-inspired score \(\hat{\Delta}_h+z\hat{\sigma}_h\) \citep{auer2002finite}, where \(z\) controls the uncertainty bonus. The leader receives the next evaluation chunk, while the alternative receives additional evaluations only if its score exceeds the acceptance margin \(\tau\). This continues until the decision is resolved or the budget is exhausted (\Cref{alg:candidate-acceptance} ). The candidate with the largest final \(\hat{\Delta}_h\) replaces the current prompt only if \(\hat{\Delta}_h>\tau\); otherwise, the current prompt is retained.

After \(T\) iterations, we select the final shared harness \(\hat h^{\mathrm{shared}}\) from the archive \(\mathcal{A}\) using the separate selection set \(\mathcal{D}_{\mathrm{sel}}\). This decouples per-iteration acceptance from final harness selection and allows an earlier candidate to be recovered if later updates generalize less well.

\subsection{\method{}-T: Evidence-Grounded Test-Time Refinement}
\label{sec:stage2}

Stage~2 of \method{} adapts the instance-specific harness while keeping the learned shared prompt fixed. Given a test instance \(x\), we first generate
\begin{equation}
    y_0
    \sim
    \pi_\theta(\cdot \mid x;\hat h^{\mathrm{shared}}).
\end{equation}
We then refine this program for at most \(B\) steps using only evidence available at test time, without access to the test solution or correctness label.

\subsubsection{Two Sources of Refinement Evidence}
Refinement draws on two complementary sources. \emph{Execution evidence} includes execution traces, runtime errors, returned outputs, and environment probes that help identify instance-specific problems. \emph{Retrieved demonstration evidence} consists of execution-verified training examples that provide implementation patterns, such as library-specific conventions, that execution feedback alone may not reveal.

The refiner combines environment interaction and execution-based program refinement \citep{yao2022react,chen2024teaching} with instance-specific demonstration retrieval. Unlike MIPROv2, which selects demonstrations during offline optimization \citep{opsahl2024optimizing}, we select demonstrations separately for each test instance. Specifically, we retrieve up to \(n\) verified training examples using lexical similarity, then let the refiner select at most three relevant demonstrations.

\subsubsection{Evidence-gated Refinement}
At refinement step \(k\), let \(e_k\) denote newly gathered execution observations and, when available, retrieved demonstrations. The instance-specific harness contains the current program and accumulated evidence:
\begin{equation}
    h_{x,k}^{\mathrm{inst}}
    =
    \left(y_{k-1}, e_{1:k}\right).
    \label{eq:instance-harness}
\end{equation}
Conditioned on this state and the fixed shared prompt, the same LLM used for initial generation proposes a revised program:
\begin{equation}
\resizebox{0.85\linewidth}{!}{$\displaystyle
    \tilde y_k
    \sim
    \pi_\theta
    \left(
        \cdot \mid
        x;
        \hat h^{\mathrm{shared}},
        h_{x,k}^{\mathrm{inst}}
    \right),
    \qquad k=1,\ldots,B.
$}
\end{equation}

We apply an evidence gate rather than accepting every revision. First, the controller executes \(\tilde y_k\) and rejects it if it fails the available programmatic checks. For revisions that pass, the same LLM compares \(y_{k-1}\) and \(\tilde y_k\) using both the accumulated evidence \(e_{1:k}\) and the revision's execution results. It assesses whether the changes address an observed issue or implement a convention supported by the retrieved demonstrations. A revision is accepted only if it passes the programmatic checks and the LLM judges its changes to be supported by the evidence.

Let \(G_k\in\{0,1\}\) denote this combined acceptance decision. We update
\begin{equation}
    y_k =
    \begin{cases}
        \tilde y_k, & G_k=1,\\
        y_{k-1}, & G_k=0.
    \end{cases}
\end{equation}
Rejected revisions leave the current program unchanged; accepted revisions become the starting point for the next step. After \(K\leq B\) refinement steps, we return \(y_K\).

\section{Experimental Setup}
\subsection{Benchmarks}
\label{sec:datasets}

We evaluate on four text-to-function generation benchmarks spanning SQL, Cypher, and Python generation.

\vspace{0.2em}
\noindent \textbf{BIRD-SQL \citep{li2023can}.}
BIRD-SQL evaluates text-to-SQL generation over realistic SQLite databases from diverse application domains. Given a natural-language question, database schema, and optional external knowledge, the model generates an executable SQL query.

\vspace{0.2em}
\noindent \textbf{CypherBench \citep{feng2025cypherbench}.}
CypherBench evaluates text-to-Cypher generation over large-scale Neo4j property graphs derived from Wikidata, covering query patterns such as multi-hop retrieval, aggregation, comparison, and temporal reasoning.

\vspace{0.2em}
\noindent \textbf{BigCodeBench-Instruct \citep{zhuo2025bigcodebench}.}
BigCodeBench-Instruct evaluates text-to-Python function generation from text instruction across diverse programming and library-use tasks.

\vspace{0.2em}
\noindent \textbf{DSCodeBench \citep{ouyang2026dscodebench}.}
DSCodeBench evaluates text-to-Python function generation for practical data-science tasks spanning ten widely used libraries.

For each benchmark, we construct a diagnostic set \(\mathcal{D}_{\mathrm{diag}}\) to provide examples and execution feedback for prompt optimization, and an update set \(\mathcal{D}_{\mathrm{upd}}\) to evaluate candidate revisions. For final harness selection, we use a separate set \(\mathcal{D}_{\mathrm{sel}}\) where dataset size permits; otherwise, we reuse \(\mathcal{D}_{\mathrm{upd}}\). Split sizes are reported in \Cref{tab:datasets}, with further details on preprocessing, split construction, and evaluation procedures in \Cref{apx:datasets}.

\begin{table*}[t]
    \centering
    \setlength{\belowcaptionskip}{6pt}
    
    \resizebox{\textwidth}{!}{%
    \begin{tabular}{lc|ccccc|cccc|c}
        \toprule
        & \multicolumn{1}{c|}{\cellcolor{gray!15}\sffamily\bfseries Baseline}
        & \multicolumn{5}{c|}{\cellcolor{gray!15}\sffamily\bfseries Offline Prompt Optimization}
        & \multicolumn{4}{c|}{\cellcolor{gray!15}\sffamily\bfseries Test-Time Adaptation}
        & \multicolumn{1}{c}{\cellcolor{gray!15}\sffamily\bfseries Harness Optimization} \\
        
        \cmidrule(lr){2-2}
        \cmidrule(lr){3-7}
        \cmidrule(lr){8-11}
        \cmidrule(lr){12-12}

        \textbf{Dataset}
        & \textbf{Seed}
        & \textbf{\method{}-O}
        & \textbf{SkillOpt}
        & \textbf{GEPA}
        & \textbf{MIPROv2}
        & \textbf{ACE}
        & \textbf{DC-Cu}
        & \makecell{\textbf{\method{}-O+T} \\ }
        & \makecell{\textbf{\method{}-O} \\ \textbf{+ ReAct}}
        & \makecell{\textbf{\method{}-O} \\ \textbf{+ DC-Cu}}
        & \makecell{\textbf{Meta-} \textbf{Harness}} \\
        \midrule

        BIRD DEV
        & 59.39 (0.32)
        & \textbf{63.55 (0.21)}
        & 61.31 (1.19)
        & 62.59 (0.59)
        & 60.87 (0.57)
        & 60.87 (0.10)
        & 60.22 (0.79)
        & \textbf{64.81} (0.21)
        & 63.12 (1.03)
        & 62.44 (1.43)
        & 58.24 (2.18) \\

        BigCodeBench
        & 47.93 (0.74)
        & \textbf{52.59 (0.31)}
        & 51.81 (0.47)
        & 50.13 (1.00)
        & 50.65 (0.59)
        & 51.96 (1.25)
        & 44.79 (0.83)
        & \textbf{53.79} (0.33)
        & 52.64 (0.24)
        & 51.70 (0.89)
        & 45.26 (2.52) \\

        DSCodeBench
        & 44.74 (0.24)
        & \textbf{54.84 (1.87)}
        & 51.87 (1.50)
        & 52.14 (1.99)
        & 51.73 (1.76)
        & 51.31 (0.43)
        & 42.81 (0.24)
        & 56.04 (1.87)
        & \textbf{56.22} (2.87)
        & 54.70 (0.87)
        & 45.99 (0.52) \\

        CypherBench
        & 74.00 (0.44)
        & \textbf{85.50 (1.40)}
        & 84.10 (1.30)
        & 78.73 (1.53)
        & 78.27 (3.04)
        & 77.13 (1.74)
        & 77.13 (1.74)
        & \textbf{89.93} (1.46)
        & 87.00 (0.21)
        & 81.77 (2.98)
        &  77.03 (2.59) \\

        \bottomrule
    \end{tabular}%
    }

    \caption{\textbf{Performance across four code-generation benchmarks.} Results (pass@1) are reported as mean (standard deviation) over three runs. Bold highlights the best mean within each category.}
    \label{tab:tool_generation_results}
    \vspace{-1.5em}
\end{table*}

\subsection{Models and Baselines}
\label{sec:baselines}

We compare against three classes of methods: \textbf{(i) offline prompt optimization}\footnote{We do not include \citet{wu2024strago} because no publicly available implementation is provided.}, including GEPA \citep{agrawal2026gepa}, MIPROv2 \citep{opsahl2024optimizing}, ACE \citep{zhang2026agentic}, and SkillOpt \citep{yang2026skillopt}; \textbf{(ii) test-time refinement}, including ReAct \citep{yao2022react} and Dynamic Cheatsheet (DC-Cu) \citep{suzgun2026dynamic}; and \textbf{(iii) end-to-end harness optimization}, represented by Meta-Harness \citep{lee2026meta}. Offline optimizers start from the same seed prompt (\Cref{apx:prompt_templates}) and use the same optimization data and execution-based correctness signal. Test-time methods use only information available at inference time, without access to reference solutions, reference outputs, or correctness labels.

For each benchmark, we set a fixed rollout budget based on GEPA's \texttt{auto=heavy} setting (\Cref{tab:datasets}) and apply the same budget to \method{} and all baselines, matching rollout counts as closely as their procedures permit. Unless otherwise stated, all methods use GPT-5 with medium reasoning effort as the backbone model. We report results over three seeds. Additional implementation details for \method{} and the baselines are provided in \Cref{apx:implementation_details}.

\subsection{Evaluation Metrics}
\label{sec:metrics}

\noindent \textbf{Execution accuracy.}
Our primary metric is pass@1. We generate one solution for each instance and report the fraction that satisfies the benchmark's correctness criteria.

\vspace{0.5em}
\noindent \textbf{Execution efficiency.}
Accuracy alone does not distinguish between correct programs with different execution costs \citep{waghjale2024ecco}. We therefore measure runtime efficiency using the Valid Efficiency Score (VES) \citep{li2023can},
\begin{equation}
\mathrm{VES}
= \tfrac{1}{N} \sum_{n=1}^{N}
\mathbf{1}(V_n=\hat V_n)
\sqrt{
\tfrac{E(Y_n)}
{E(\hat Y_n)}
},
\end{equation}
where $V_n$ and $\hat V_n$ denote the reference and predicted execution results, and $E(Y_n)$ and $E(\hat Y_n)$ their corresponding execution times. Incorrect solutions receive zero credit, while faster correct solutions receive higher scores. Detailed timing and scoring protocols are provided in \Cref{apx:ves}.

\vspace{0.5em}
\noindent \textbf{Cumulative fix-to-break ratio.}
To characterize how reliably performance improves during offline optimization, we track whether accepted prompt updates fix previously incorrect instances or break previously correct ones. At optimization progress $t$, we define
\begin{equation}\label{eq:flip_ratio}
\rho(t)
=
\tfrac{\sum_{s=0}^{t}\mathrm{fixed}_s}
     {\sum_{s=0}^{t}\mathrm{broken}_s},
\end{equation}
where \(\mathrm{fixed}_s\) and \(\mathrm{broken}_s\) count incorrect-to-correct and correct-to-incorrect transitions induced by accepted updates at iteration \(s\). A ratio \(\rho(t)>1\) indicates that accepted updates cumulatively correct more instances than they disrupt.
\section{Results}
\subsection{Fixes and Regressions During Prompt Optimization}
\label{sec:result_flip}
We first examine how offline prompt optimization affects program generation. While validation accuracy captures the net effect of an update, it does not reveal the balance between fixing incorrect instances and introducing regressions on previously correct ones. We therefore analyze the cumulative fix-to-break ratio (\Cref{eq:flip_ratio}) to characterize this balance over optimization.

As shown in \Cref{fig:flip_ratio}, \method{}-O maintains more fixes than regressions throughout optimization and achieves the highest ratios overall. On BIRD-SQL, \method{}-O reaches a cumulative ratio of approximately \(3.0\), compared with \(2.0\) for MIPROv2 and SkillOpt, \(1.5\) for GEPA, and \(1.2\) for ACE. ACE, which incrementally accumulates instance-derived strategy bullets, initially favors fixes but gradually approaches a ratio of \(1.0\), whereas methods that make broader prompt revisions maintain higher ratios. A similar pattern holds on BigCodeBench, where \method{}-O reaches approximately \(2.2\), comparable to GEPA and higher than MIPROv2 (\(1.4\)) and ACE (\(1.2\)). Our ablation further shows that using only failed cases reduces the average fix-to-break ratio from \(2.23\) to \(1.91\) (\Cref{tab:flip-ratio-ablation}), suggesting that combining successful and failed cases provides more informative feedback for prompt revisions than failures alone. These results suggest that our feedback design helps accumulate fixes while limiting regressions during prompt optimization.

\begin{figure}[t!]
    \centering
    \includegraphics[width=1.0\linewidth]{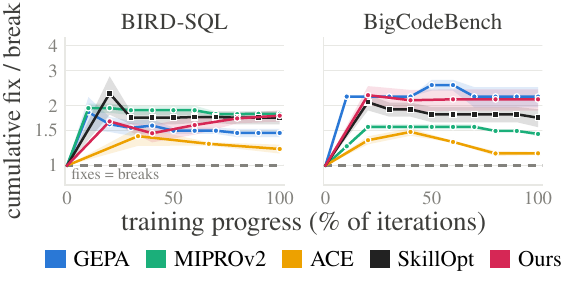}
    \vspace{-2.0em}
    \caption{\textbf{Cumulative fix-to-break ratio.} The y-axis reports the cumulative ratio across optimization iterations on the x-axis. The dashed line indicates an equal number of fixes and regressions.}
    \label{fig:flip_ratio}
\vspace{-1.0em}
\end{figure}

\vspace{-0.5em}
\subsection{Benchmark Performance of Offline Prompt Optimization}
\label{sec:main_results}

We next evaluate whether the resulting prompts improve pass@1 on held-out text-to-SQL, text-to-Cypher, and text-to-Python code-generation tasks.

As shown in \Cref{tab:tool_generation_results}, \method{}-O improves pass@1 over the initial prompt across all four benchmarks, with gains of \(4.16\) points on BIRD-SQL, \(4.66\) on BigCodeBench, \(10.1\) on DSCodeBench, and \(11.5\) on CypherBench. \method{}-O also achieves the highest mean pass@1 among offline prompt optimizers on all four benchmarks. It reaches \(63.5\) on BIRD-SQL, compared with \(62.5\) for GEPA and \(61.3\) for SkillOpt; \(52.5\) on BigCodeBench, compared with \(51.9\) for ACE and \(51.8\) for SkillOpt; \(54.8\) on DSCodeBench, compared with \(52.1\) for GEPA; and \(85.5\) on CypherBench, compared with \(84.1\) for SkillOpt. By contrast, Meta-Harness, which optimizes the end-to-end harness, does not consistently improve performance and can underperform the initial prompt, achieving \(58.2\) on BIRD-SQL and \(45.3\) on BigCodeBench.

\begin{figure}[h!]
    \centering
    \includegraphics[width=1.0\linewidth]{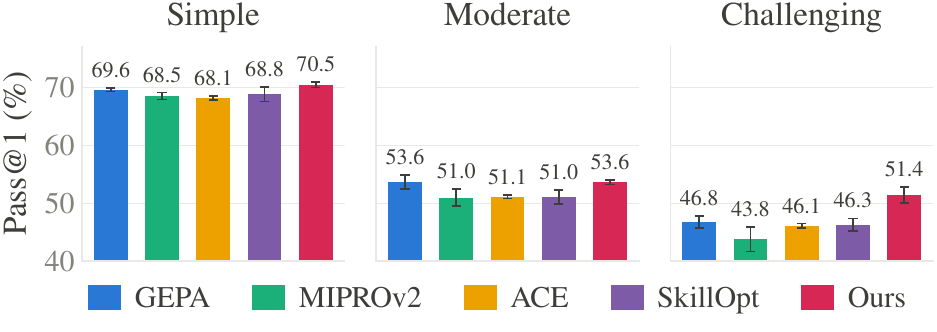}
    \vspace{-2.0em}
    \caption{BIRD-SQL DEV pass@1 by question difficulty (simple $n{=}923$, moderate $n{=}463$, challenging $n{=}144$ scored instances). Bars are the mean over three seeds; error bars are the sample standard deviation.}
    \label{fig:bird_result}
    \vspace{-0.3em}
\end{figure}

We further examine performance across pre-specified BIRD-SQL difficulty levels (\Cref{fig:bird_result}). \method{}-O achieves pass@1 scores of \(70.5\), \(53.6\), and \(51.4\) on simple, moderate, and challenging questions, respectively. Its advantage is largest on challenging questions, where \method{}-O reaches \(51.4\), outperforming the next-best method, GEPA (\(46.8\)), by \(4.6\) points. These results show that the gains from grounded prompt optimization persist across difficulty levels and are most pronounced on challenging instances.

\paragraph{Learned prompts and optimization cost.}
\begin{table}[t!]
\centering
\small
\resizebox{\columnwidth}{!}{%
\begin{tabular}{lcc}
\toprule
\textbf{Method}
& \textbf{Tokens/call (K)}
& \textbf{Prompt length (tokens)} \\
\midrule
ACE      & 31.89 (1.17) & 35,881 (7,563) \\
MIPROv2  & 3.47 (0.02) & 2,704 (70) \\
SkillOpt & 2.87 (0.10) & 2,021 (628) \\
GEPA     & \textbf{2.47} (0.24) & 2,003 (541)\\
\method{}-O (Ours) & 2.79 (0.18)  & \textbf{1,949} (389)\\
\bottomrule
\end{tabular}%
}
\caption{\textbf{Optimization context and token cost.} Input plus output tokens per LLM call during optimization and the length of the final prompt. Averaged over four benchmarks per seed; mean (SD) over three seeds.}

\label{tab:optimization_cost}
\vspace{-1.5em}
\end{table}
We next examine whether these performance gains come at the cost of larger prompts or greater per-call token usage (\Cref{tab:optimization_cost}). \method{}-O, GEPA, and SkillOpt produce compact instruction sets, whereas MIPROv2 additionally selects few-shot demonstrations and ACE accumulates a substantially larger playbook. Averaged across benchmarks, \method{}-O produces the shortest final prompts (1,949 tokens), compared with 2,003 for GEPA, 2,704 for MIPROv2, and 35,881 for ACE. Our cost analysis also shows that more compact prompts generally correspond to lower per-call token usage: \method{}-O averages 2.79K tokens per optimization call, comparable to GEPA (2.47K) and SkillOpt (2.87K), and substantially below ACE (31.89K).

Upon inspection of the learned prompt, we found that \method{}-O primarily retains general task-level instructions (\Cref{fig:our-prompt-bird}), whereas other optimizers preserve varying amounts of training-specific content (\Cref{fig:ace-prompt,fig:gepa_optimized-prompt-bird}). Together, these results suggest that \method{} learns compact, reusable guidance rather than accumulating instance-specific details, while achieving its performance gains without substantially larger prompts (\Cref{fig:prompt_length_vs_cost}).

\subsection{Test-Time Refinement Complements Offline Prompt Optimization}
\label{sec:result_refinement}

Offline prompt optimization captures reusable task-level guidance but may leave instance-specific errors unresolved. We therefore evaluate whether \method{}-T---evidence-grounded test-time refinement---provides additional gains and how it compares with alternative test-time adaptation strategies. For each dataset, we initialize test-time refinement with the best offline prompt obtained from \method{}-O.

As shown in \Cref{tab:tool_generation_results}, \method{}-T improves pass@1 on all benchmarks , by \(1.26\) points on BIRD-SQL, \(1.20\) on BigCodeBench, \(1.20\) on DSCodeBench, and \(4.43\) on CypherBench, for an average gain of \(2.02\) points. The resulting system achieves the highest mean pass@1 on three of four benchmarks; on DSCodeBench, ReAct is marginally higher (\(56.22\) vs.\ \(56.04\)). Alternative test-time methods are less consistent: ReAct improves performance on three benchmarks but slightly degrades BIRD-SQL, while DC-Cu decreases performance on all four when applied after offline optimization. To distinguish whether this behavior arises from combining DC-Cu with offline optimization, we also apply DC-Cu directly to the initial prompt. The results remain mixed: it improves BIRD-SQL and CypherBench by \(0.83\) and \(3.13\) points, respectively, but decreases BigCodeBench and DSCodeBench by \(3.14\) and \(1.93\) points.

These results suggest that the two stages of \method{} are complementary: offline optimization captures reusable task-level guidance, while test-time refinement provides additional improvements for individual instances.

\vspace{-0.5em}
\subsection{Execution Efficiency}
\label{sec:result_efficiency}

\begin{table}[t!]
    \centering
    \setlength{\belowcaptionskip}{6pt}
    \resizebox{\columnwidth}{!}{%
    \begin{tabular}{lcccc}
        \toprule
        \textbf{Method}
        & \textbf{BIRD}
        & \textbf{BigCodeBench}
        & \textbf{DSCodeBench}
        & \textbf{CypherBench} \\
        \midrule

        Seed prompt
        & 79.1 (5.0)
        & 61.2 (3.6)
        & 48.7 (0.5)
        & 85.6 (1.5) \\

        \midrule
        \multicolumn{5}{c}{\cellcolor{gray!15}\sffamily\bfseries Stage Offline Harness Optimization} \\

        GEPA
        & 73.1 (3.6)
        & 69.1 (3.9)
        & 56.0 (2.2)
        & 86.7 (3.8) \\

        MIPROv2
        & 77.9 (7.1)
        & 69.7 (2.7)
        & 55.4 (1.3)
        & 86.5 (3.2) \\

        ACE
        & 72.3 (2.2)
        & 71.3 (2.5)
        & 55.3 (0.5)
        & 103.8 (15.2) \\

        SkillOpt
        & 75.9 (4.0)
        & 69.7 (1.0)
        & 54.8 (1.0)
        & \textbf{137.7 (50.8)} \\
        Metaharness 
        & 74.5 (10.3)
        & 58.5 (3.0)
        & 84.8 (4.3) 
        & 50.5 (0.7) \\
        {\method{}-O}
        & \textbf{91.9 (6.4)}
        & \textbf{75.0 (3.5)}
        & \textbf{59.8 (1.7)}
        & 109.8 (39.8) \\

        \midrule
        \multicolumn{5}{c}{\cellcolor{gray!15}\sffamily\bfseries Stage Test-Time Adaption} \\
        {DC-Cu}
        & 83.7 (4.5)
        & 54.5 (7.2)
        & 48.8 (0.8)
        & 74.2 (1.9) \\
        \method{}-O + DC-Cu
        & 91.4 (4.9)
        &  70.6 (1.9)
        &  60.0 (1.2)
        & 92.6 (16.0) \\
        {\method{}-O + ReAct}
        & 92.3 (7.6)
        & 76.2 (3.6)
        & 60.7 (2.3)
        & 110.1 (42.8) \\
        {\method{}-O + T}
        & \textbf{93.4 (6.1)}
        & \textbf{76.2 (3.6)}
        & \textbf{61.5 (1.4)}
        & 101.7 (7.8) \\

        \bottomrule
    \end{tabular}%
    }
    \caption{\textbf{Valid Efficiency Score (VES) of generated programs.} Results are mean (SD) over three seeds; higher is better. Bold denotes the best mean.}    \label{tab:ves}
    \vspace{-2em}
\end{table}

For SQL queries and Python code, correctness alone does not fully capture solution quality: programs that produce the same correct result can differ substantially in execution cost \citep{li2023can}. We therefore report VES alongside pass@1 to assess whether improvements in correctness also preserve execution efficiency.

As shown in \Cref{tab:ves}, \method{}-O achieves the highest mean VES on three of four benchmarks: BIRD-SQL (\(91.9\)), BigCodeBench (\(75.0\)), and DSCodeBench (\(57.8\)). Together with the corresponding pass@1 gains, these results indicate that the improved correctness on these tasks is not achieved at the expense of execution efficiency. On CypherBench, SkillOpt attains the highest mean VES (\(137.7\)), though with substantial variability across runs (SD \(50.8\)). With \method{}-T, VES further improves on BIRD-SQL, BigCodeBench, and DSCodeBench, alongside the corresponding gains in pass@1. On CypherBench, however, pass@1 increases from \(85.5\) to \(89.93\), while VES decreases from \(109.8\) to \(101.7\). This divergence highlights a correctness--efficiency trade-off: test-time refinement solves more instances, but the resulting programs are less efficient under VES. Overall, \method{} delivers concurrent gains in correctness and execution efficiency on three of four benchmarks, demonstrating that its performance improvements generally extend beyond accuracy alone.

\subsection{Ablation Study and Sensitivity Analysis}
\label{sec:ablation}

We ablate the main components of offline optimization on BIRD-SQL and BigCodeBench and examine sensitivity to the amount of diagnostic evidence and optimization budget.

\begin{figure}[h!]
    \centering
    \includegraphics[width=0.95\linewidth]{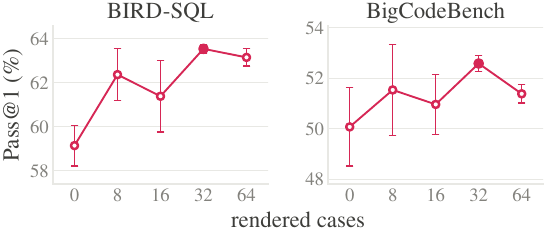}
    \vspace{-0.5em}
    \caption{\textbf{Sensitivity to the number of rendered diagnostic cases.} Pass@1 as the number of cases rendered varies. Results are the mean $\pm$ SD over three seeds.}    
    \label{fig:sensitivity}
    \vspace{-1.0em}
\end{figure}

\paragraph{Ablation study.} Removing the rendered success and failure cases consistently reduces performance: pass@1 drops from \(63.55\) to \(59.15\) on BIRD-SQL and from \(52.5\) to \(50.08\) on BigCodeBench (\Cref{fig:sensitivity}). This indicates that validation signals alone do not provide sufficient guidance for prompt revision; concrete execution examples help the optimizer identify what to correct and what to preserve.

\paragraph{Sensitivity analysis and model generalization.}
\method{} remains relatively stable across a range of configurations, including the number of candidates, optimization rounds, and rendered diagnostic examples, with performance consistently above the seed prompt. Similarly, increasing the test-time refinement budget yields limited additional gains beyond the default $B=3$ (\Cref{apx:refinement_budgets}). These results suggest that the observed improvements are not highly sensitive to hyperparameter choices or prolonged refinement. Experiments with GPT-5.4 (\Cref{tab:gpt54_results}) further show consistent gains over the evaluated baselines, indicating that the approach extends beyond the GPT-5 backbone.



\vspace{-0.5em}
\section{Conclusion}
In this work, we introduced \method{}, a two-stage harness framework for program generation that combines grounded offline prompt optimization with evidence-grounded test-time refinement. Across four SQL, Cypher, and Python benchmarks, \method{} consistently outperforms the initial prompt and existing prompt optimization baselines, with test-time refinement providing further gains on every benchmark. Our analyses show that execution-grounded updates reduce regressions during optimization, while the learned prompts remain compact and robust across optimization settings. Together, these results support combining reusable cross-task optimization with instance-specific refinement for more effective program generation.

\section*{Limitations}
\paragraph{Budgeting and computational cost.}
Our evaluation uses fixed, benchmark-specific budgets, which may limit the gains that some methods could achieve with longer searches. Substantially expanding these budgets is computationally demanding: even under the evaluated budgets, ACE and Meta-Harness require approximately 12 and 24 hours of wall-clock time, respectively, on BigCodeBench in our setup. To support controlled comparisons within these practical constraints, we use task-rollout counts as a common measure of evaluation effort and match budgets as closely as method-specific procedures permit. This measure, however, does not fully capture differences in the costs of reflection, candidate generation, and selection. Nevertheless, we have sought to make the comparisons as fair and transparent as possible within these computational and methodological constraints.


\paragraph{Model dependency.}
Our main evaluation relies on proprietary GPT-5 and GPT-5.4 models, whose future updates or deprecation may affect exact harness reproducibility. However, \method{} operates entirely at the harness level and does not require access to or modification of model weights, allowing the same optimization procedure to be applied to other compatible backbones. Learned prompts may still require re-optimization when the underlying model changes.



\bibliography{custom}

\clearpage
\appendix
\section*{Appendix}\label{sec:appendix}
\section{Datasets Details}\label{apx:datasets}
\begin{table}[t!]
\centering
\small
\resizebox{\linewidth}{!}{%
\begin{tabular}{@{}llrrrrr@{}}
\toprule
Dataset & Task
& $\mathcal{D}_{\mathrm{diag}}$
& $\mathcal{D}_{\mathrm{upd}}$
& $\mathcal{D}_{\mathrm{sel}}$
& $\mathcal{D}_{\mathrm{test}}$
& Rollout budget \\
\midrule
BIRD-SQL     & Text-to-SQL    & 500 & 250 & 1,400 & 1,530 & 5,400 \\
CypherBench  & Text-to-Cypher & 250 & 250 & 250   & 1,000 & 3,700 \\
BigCodeBench & Python         & 250 & 250 & 250   & 637   & 5,400 \\
DSCodeBench  & Python         & 250 & 250 & 250   & 482   & 3,600 \\
\bottomrule
\end{tabular}%
}
\caption{\textbf{Benchmarks, data splits, and optimization budgets.}
Rollout budgets cap task executions during offline optimization.}
\label{tab:datasets}
\vspace{-1.5em}
\end{table}
Here, we provide additional details on each benchmark and our preprocessing and splitting procedures. Three conventions apply across all datasets. First, all splits are constructed once using a fixed dataset-level seed before any experiment, and pool membership is held fixed across methods and run seeds. Second, the diagnosis pool $\mathcal{D}_{\mathrm{diag}}$, update pool $\mathcal{D}_{\mathrm{upd}}$, and, when applicable, selection pool $\mathcal{D}_{\mathrm{sel}}$ are disjoint at the instance level. Third, the evaluation split is never accessed during optimization and is used only for final evaluation of the selected prompt.

\paragraph{BIRD-SQL \citep{li2023can}.}

BIRD-SQL is a large-scale text-to-SQL benchmark designed to evaluate SQL generation in realistic database settings, where models must reason over database contents in addition to schemas. It contains 12,751 natural-language--SQL pairs across 95 SQLite databases and 37 application domains: 9,428 training examples over 69 databases, 1,534 development examples over 11 databases, and 1,789 examples over 15 databases in a concealed test set. Among the training and development databases, 32\% originate from Kaggle, 48\% from the CTU Prague Relational Learning Repository, and the remaining 20\% from other publicly available tables. Each instance includes a natural-language question, a reference SQL query, the associated database, and optional expert-annotated external knowledge. The development set spans three difficulty levels: 925 simple, 465 moderate, and 144 challenging examples, categorized based on question understanding, external knowledge requirements, database complexity, and SQL complexity. BIRD also evaluates execution efficiency, enabling assessment of both correctness and runtime behavior. 

\paragraph{Preprocessing and splits.}
Following StreamBench \citep{wu2024streambench}, we use the official TRAIN split for optimization and the official DEV split for evaluation, both downloaded from \footnote{\url{https://bird-bench.github.io/}}; the concealed test set is not used. To evaluate cross-database transfer, we partition TRAIN at the database level. We reserve 14 of the 69 training databases for final prompt selection, sampling at most 100 questions per database. Of the remaining 55 databases, 15 are reserved exclusively for the update pool, from which we sample 250 questions, while the 500-question diagnosis pool is sampled from the remaining 40 databases with every database represented. Thus, the diagnosis, update, and selection pools are disjoint at the database level.

Before sampling, we exclude examples whose reference queries fail to execute or return empty results. For final evaluation, we use the 1,534 DEV examples and exclude four whose reference queries fail under our evaluator, yielding 1,530 scored examples. Correctness is determined using the official execution-accuracy comparator: predicted and reference outputs are compared as sets of tuples, ignoring row order and duplicates while preserving column order. 


\paragraph{BigCodeBench-Instruct \citep{zhuo2025bigcodebench}.} BigCodeBench evaluates Python function generation on practical programming tasks requiring complex instruction following and composition of multiple library APIs. The benchmark contains 1,140 function-level tasks spanning seven application domains and 139 libraries, covering 723 distinct function calls. Tasks are derived from ODEX examples originating from Stack Overflow and are subsequently refactored, tested, and manually curated through a human--LLM collaborative process. Each task includes an executable reference implementation and unit tests, with an average of 5.6 tests and 99\% branch coverage. BigCodeBench provides two prompt variants over the same tasks: \emph{BigCodeBench-Complete}, which uses structured docstrings, and \emph{BigCodeBench-Instruct}, which converts them into concise natural-language instructions. We use the BigCodeBench-Instruct release from \footnote{\url{https://huggingface.co/datasets/bigcode/bigcodebench}}. 

\paragraph{Preprocessing and splits.} Because BigCodeBench does not provide an official training split, we first audit all reference solutions in the official Docker environment. We exclude three tasks whose reference solutions do not pass their own tests: tasks 590 and 1005 depend on live network endpoints, while task 461 exceeds the canonical execution-time limit. We then shuffle the remaining 1,137 tasks once using a fixed seed. The first 500 form the optimization pool, divided equally into 250 diagnosis and 250 update examples; the remaining 637 examples form a frozen evaluation set. No separate selection pool is used, so final prompt selection is performed on the update pool.

The model is given the benchmark's \texttt{instruct\_prompt} as input and generates a single code block. We evaluate generated programs using the official execution protocol and Docker environment with the benchmark's hidden unit tests. We additionally apply the official post-processing procedure to extract valid function implementations and remove code unrelated to the target entry point. During optimization, the model receives only pass/fail feedback and exception types; assertion messages containing expected outputs are withheld. Because the model endpoint does not support explicit temperature control, pass@1 is computed from a single generation using the default decoding configuration. For stratified sampling, tasks are grouped by their primary library into ten groups.

\paragraph{CypherBench \citep{feng2025cypherbench}.}
CypherBench evaluates natural-language-to-Cypher generation over large-scale Neo4j property graphs. It is designed to assess whether LLMs can query realistic knowledge graphs without relying on simplified schemas or external entity-linking pipelines. The benchmark contains 11 property graphs derived from an April 2024 Wikidata snapshot, comprising approximately 7.8 million entities across domains including art, biology, sports, companies, geography, movies, and politics. Questions cover 12 graph-query patterns, including multi-hop reasoning, aggregation, comparison, temporal reasoning, grouping, and global retrieval, and are constructed using templates followed by LLM-based rewriting and verification. We use the released CypherBench dataset from Hugging Face,\footnote{\url{https://huggingface.co/datasets/megagonlabs/cypherbench}} which contains 10,882 question--Cypher pairs across 11 property graphs.

\paragraph{Preprocessing and splits.}
Rather than using the original graph-level train/test partition, we combine all 10,882 released questions and construct fixed question-level splits stratified by graph and query pattern. Using proportional sampling, we select 250 examples for the diagnosis pool, 250 for the update pool, and 1,000 for evaluation; the remaining examples are unused. We do not use a separate selection pool, and final prompt selection is performed on the update pool. Because the same graphs may appear in both the optimization and evaluation sets, this setting evaluates performance on previously observed graphs rather than generalization to unseen graphs.

Before constructing the splits, we execute all reference queries against the corresponding Neo4j graphs. All queries execute successfully within the 120\,s timeout and return non-empty results, so no examples are excluded. The model is given the graph schema in the official \texttt{PropertyGraphSchema} format together with the natural-language question and generates a Cypher query. Generated queries are executed in a read-only Neo4j session. Following the benchmark evaluation protocol, predicted and reference outputs are compared as multisets while allowing column permutations. Row order is considered only when the reference query explicitly specifies an ordering, and queries that result in a Neo4j execution error are marked incorrect. We use a 60\,s query timeout during optimization and a 120\,s timeout during final evaluation. For stratified sampling and analysis, examples are grouped by graph.

\paragraph{DSCodeBench \citep{ouyang2026dscodebench}.}
DSCodeBench evaluates realistic data-science code generation across ten Python libraries. It contains 1,000 function-generation problems covering NumPy (131), Pandas (92), SciPy (112), Scikit-learn (108), Matplotlib (105), Seaborn (83), TensorFlow (110), PyTorch (101), Keras (104), and LightGBM (54). The benchmark is constructed using examples from DS-1000 and highly rated Stack Overflow answers to retrieve realistic implementations from public GitHub repositories. Retrieved code is converted into self-contained functions, filtered and perturbed to reduce direct code reuse and potential data leakage, and paired with natural-language problem descriptions and execution-based test cases, followed by manual verification. Each problem is evaluated using 200 test cases. We use the dataset released at \footnote{\url{https://github.com/ShuyinOuyang/DSCodeBench}}.

\paragraph{Preprocessing and splits.}
Because DSCodeBench does not provide an official training split, we first validate all reference solutions in the benchmark's pinned CPU environment. Each reference solution is executed twice on the official 200 seeded inputs and evaluated using the benchmark's official comparator. We exclude 18 problems: seven whose reference solutions are incompatible with our evaluation setting because they fail under the official comparator, require GPU-only operations, or exceed a 30-minute timeout, and eleven that produce inconsistent outputs across repeated executions. We randomly shuffle the remaining 982 problems using a fixed seed. The first 500 form the optimization set, split evenly into 250 diagnosis and 250 update examples, while the remaining 482 form the evaluation set. We do not use a separate selection set; final prompt selection is performed on the update set.

The model is given the benchmark's problem description as input and generates Python code. Following the official evaluation harness, we extract the first generated code block and invoke its final top-level function. A prediction is considered correct only if its outputs match the reference outputs on all 200 seeded inputs according to the benchmark's type-specific comparator, with a 200-second execution limit per candidate. Reference outputs are precomputed and reused across candidate solutions. The test script is not provided to the model, and evaluation messages containing expected outputs are withheld during optimization. We use the same initial prompt as for BigCodeBench.

\subsection*{Software Environment}
Benchmark-specific execution environments follow the official implementations where available. BIRD-SQL queries are executed with SQLite 3.53.4 through Python's \texttt{sqlite3} module.\footnote{\url{https://docs.python.org/3/library/sqlite3.html}} BigCodeBench programs are evaluated using the official Docker image (v0.2.1.post7),\footnote{\url{https://hub.docker.com/r/bigcodebench/bigcodebench-evaluate/tags}} including the official code post-processing procedure. CypherBench queries are executed against Neo4j 5.20.0 Community Edition\footnote{\url{https://neo4j.com/}} using the \texttt{neo4j} Python driver v6.3.1\footnote{\url{https://pypi.org/project/neo4j/6.3.1/}} in read-only sessions. DSCodeBench programs are evaluated on CPU using the benchmark's pinned Python environment.\footnote{\url{https://github.com/ShuyinOuyang/DSCodeBench}}

\section{Baseline Methods}\label{apx:baselines}
We compare against methods that optimize different components of the harness defined in \Cref{eq:joint_harness}. Offline prompt optimization methods modify the shared harness \(h^{\mathrm{shared}}\) using optimization data; test-time methods adapt external state during inference while keeping the model parameters fixed; and Meta-Harness searches over a broader shared harness that may include both prompts and executable control logic. Unless otherwise stated, all methods use the same initial prompt, task interface, and execution-based evaluator as our method.

\subsection{Offline Prompt Optimization}

Offline prompt optimizers modify only the shared harness while leaving instance-specific state unchanged. Given optimization data \(\mathcal D\), they can be viewed as searching for
\begin{equation}
    \begin{aligned}
    \hat h^{\mathrm{shared}}
    &=
    \arg\max_{h^{\mathrm{shared}}\in\mathcal H}
    \frac{1}{|\mathcal D|}
    \sum_{x\in\mathcal D}
    \\[-0.25em]
    &\qquad
    \mathbb{E}_{y\sim\pi_\theta(\cdot\mid x;h^{\mathrm{shared}})}
    \left[R(y,x)\right].
    \end{aligned}
\end{equation}
with different methods defining different representations and update procedures for \(h^{\mathrm{shared}}\).

\paragraph{GEPA \citep{agrawal2026gepa}.}
GEPA is a reflective prompt optimizer that uses execution trajectories and evaluator feedback to iteratively improve the shared system prompt. An LLM reflects on the behavior of existing prompt candidates, identifies failure modes, and proposes targeted modifications. Candidate prompts are evaluated on task examples, and a Pareto frontier over per-instance performance guides subsequent proposal and selection. In our formulation, GEPA optimizes \(h^{\mathrm{shared}}\) with the system prompt as its only mutable component.

\vspace{0.5em}
\noindent \textbf{Implementation details.} We use \texttt{gepa} package (v0.1.1)\footnote{\url{https://github.com/gepa-ai/gepa}} with its default configuration, including Pareto-based selection, reflection minibatches of three, and no prompt merging. The system prompt is initialized from our common seed prompt. The diagnosis and update pools serve as GEPA's training and validation sets, respectively, and reflection records contain the execution verdict, failure category, and evaluator feedback. We evaluate 21 candidates on BIRD-SQL and 20 on BigCodeBench; for CypherBench and DSCodeBench, we use metric-call budgets of 3,700 and 3,600, respectively.

\paragraph{MIPROv2 \citep{opsahl2024optimizing}.}
MIPROv2 optimizes a shared prompt configuration consisting of an instruction and a set of few-shot demonstrations. It generates candidate instructions from task and data summaries and uses Bayesian optimization to search over instruction--demonstration combinations. In our formulation, \(h^{\mathrm{shared}}\) comprises the optimized instruction and selected demonstrations.

\vspace{0.5em}
\noindent \textbf{Implementation details.} We adapt the reference \texttt{mipro\_optimizer\_v2} implementation from DSPy\footnote{\url{https://github.com/stanfordnlp/dspy}} to our setting using Optuna v5.0.0\footnote{\url{https://github.com/optuna/optuna}} with a TPE sampler. We retain the three-stage optimization procedure, using 18 candidate instructions and 18 demonstration sets with up to four demonstrations each. Demonstrations are bootstrapped using the initial program and supplemented with labeled examples when necessary. Candidate configurations are evaluated on minibatches of 35 examples, with a full evaluation on the update pool every five trials. We use metric-call budgets of 5,400 for BIRD-SQL and BigCodeBench, 3,700 for CypherBench, and 3,600 for DSCodeBench. The final shared harness consists of the selected instruction and its four demonstrations.

\paragraph{ACE \citep{zhang2026agentic}.}
Agentic Context Engineering (ACE) represents the shared harness as an evolving playbook of reusable strategies. Its Generator--Reflector--Curator procedure solves training tasks, extracts lessons from execution trajectories and evaluator feedback, and incrementally incorporates these lessons into the playbook. In our formulation, the learned playbook augments the initial prompt as part of \(h^{\mathrm{shared}}\).

\vspace{0.5em}
\noindent \textbf{Implementation details.}
We use the authors' offline batched implementation,\footnote{\url{https://github.com/ace-agent/ace}} retaining the original prompts and seven-section playbook format. We use three epochs with up to three Reflector rounds per epoch on BIRD-SQL, and five epochs with up to five Reflector rounds per epoch on BigCodeBench, CypherBench, and DSCodeBench. The diagnosis pool is processed in batches of 25, with the Curator constructing a delta context after every 10 reflections and incrementally updating the playbook. The token budget is treated as a soft constraint. For BIRD-SQL, we select among intermediate playbook snapshots using the held-out selection set; for the remaining benchmarks, we use the final playbook.

\subsection{Test-Time Adaptation}
Test-time methods modify external state during inference while keeping the learned shared prompt and model parameters fixed. For methods that refine the current instance, we write the update generically as
\begin{equation}
    h_{x,t+1}^{\mathrm{inst}}
    =
    \mathcal U
    \left(
        h_{x,t}^{\mathrm{inst}},
        x,
        e_t
    \right),
\end{equation}
where \(e_t\) denotes evidence obtained from the current interaction with the execution environment.

\paragraph{Dynamic Cheatsheet (DC-Cu) \citep{suzgun2026dynamic}.}
Dynamic Cheatsheet maintains a textual memory that is updated during inference. We use its \texttt{cumulative} mode, DC-Cu, in which the model generates a solution conditioned on the current memory and subsequently updates the memory based on the problem and generated output. Because this memory persists across evaluation examples, it acts as an online-updated component of \(h^{\mathrm{shared}}\).

\vspace{0.5em}
\noindent \textbf{Implementation details.}
We use the authors' reference implementation,\footnote{\url{https://github.com/suzgunmirac/dynamic-cheatsheet}} adapting the generator instructions to each task family while retaining the original curator prompt. For each example, the model may execute its generated program or query up to three times before the memory is updated. The update uses only the problem and model-generated outputs; no reference solutions, correctness labels, or evaluator verdicts are provided. The current memory and task-specific generator instructions are included in the model context for subsequent examples. We limit the memory to 20,000 characters and maintain five independent memory streams over the evaluation set to reduce sensitivity to example ordering.

\paragraph{ReAct \citep{yao2022react}.}
ReAct interleaves language-model reasoning with actions in an external environment. We use ReAct as an instance-level refinement baseline: the shared harness \(h^{\mathrm{shared}}\) remains fixed, while the current program and interaction history constitute \(h_x^{\mathrm{inst}}\). Execution feedback from the current instance is used to iteratively revise the generated program.

\vspace{0.5em}
\noindent \textbf{Implementation details.}
Because the original ReAct implementation\footnote{\url{https://github.com/ysymyth/ReAct}} uses free-form prompting for text-based tasks, we adapt ReAct to our test-time interaction framework. An initial candidate is first generated using the fixed shared prompt. During refinement, the model can invoke three tools through native function calling: a read-only diagnostic probe, execution of the current candidate, and submission of the final solution. We allow up to \(B=3\) environment interactions, excluding the final submission, and provide no few-shot ReAct trajectories. Model reasoning remains internal rather than being emitted as explicit thought--action traces.

\subsection{End-to-End Harness Optimization}

\paragraph{Meta-Harness \citep{lee2026meta}.}
Meta-Harness optimizes a broader shared harness than prompt-only methods. An agentic proposer generates candidate harness implementations using information from previously evaluated candidates, including their source code, execution traces, and validation performance. In our formulation, Meta-Harness optimizes \(h^{\mathrm{shared}}\) over a search space that includes prompting, memory, multi-call control flow, and self-verification, while keeping the underlying policy \(\pi_\theta\) fixed.

\vspace{0.5em}
\noindent \textbf{Implementation details.}
We build on the authors' reference implementation,\footnote{\url{https://github.com/stanford-iris-lab/meta-harness}} retaining its optimization loop and Pareto-frontier selection procedure while adapting the data interface to our four benchmarks. We use GPT-5 as the proposer through the OpenAI Codex CLI and generate three candidate harnesses per iteration using the original skill instructions. Candidate harnesses may use read-only diagnostic tools, while evaluation examples and reference labels are not accessible during optimization. Each candidate is optimized on 100 diagnosis examples for BIRD-SQL and BigCodeBench and 50 for CypherBench and DSCodeBench, and evaluated on the full update pool. Search terminates when the corresponding rollout budget is reached, requiring 4--11 iterations across benchmarks. The final harness is selected based on update-pool performance, with the initial harness retained when no candidate achieves better performance.

\subsection{Software environment.}
All optimization, baseline, and evaluation pipelines run under Python 3.12.14. We use GPT-5 (snapshot \texttt{2025-08-07}) as the generator for all methods through Azure OpenAI. Generation uses a maximum of 16,384 completion tokens and the deployment's default sampling and reasoning settings. Additional baseline-specific dependencies and configurations are described in \Cref{apx:baselines}.

\section{Implementation Details of \method}\label{apx:implementation_details}
\paragraph{Model configuration.}
We use GPT-5 (snapshot \texttt{2025-08-07}) for program generation, prompt optimization, and test-time refinement, with a maximum of 16,384 completion tokens and the deployment's default sampling and reasoning settings.

\subsection{Offline Prompt Optimization}

We run \(T=5\) optimization rounds with no early stopping. In each round, we evaluate the current prompt on the full diagnosis set and uniformly sample 32 incorrect and 32 correct examples. For incorrect examples, the reflector receives the task input, generated output, failure category, evaluator feedback, and available reference information; correct examples are included to identify behaviors that should be preserved. The optimization context also includes aggregate failure statistics and complementary feedback from the previous round. Based on this information, the reflector independently proposes four candidate prompt revisions, with up to eight edits per candidate. Proposals are rejected if they cannot be parsed, leave the prompt unchanged, or introduce task-specific values or identifiers found only in reference solutions.

\paragraph{Candidate evaluation and acceptance.}
In each optimization round, we compare four candidate prompts with the current prompt using adaptive paired evaluation (\Cref{alg:candidate-acceptance}). All candidates are initially evaluated on the same 50 examples from the 250-example update set, with a budget of two additional 50-example pulls allocated by the algorithm. Candidate scores are paired improvements in execution accuracy, macro-averaged across observed databases or domains. Standard errors are computed from the sample standard deviation of group means divided by the square root of the number of observed groups, floored by the pooled row-level standard error with minimum per-row variance \(0.02\). We set \(z=1\) for evaluation allocation and early stopping, but accept an update only if the best candidate's estimated mean improvement exceeds \(\tau=0.01\). Accepted prompts and any additional candidates exceeding this threshold are archived for final selection.

\paragraph{Final selection.}
After five optimization rounds, we select the final prompt through a second adaptive race over the archive, with a budget equal to \(75\%\) of the evaluations needed to score every archived prompt on the full selection set.

\subsection{Test-Time Refinement}
\label{apx:refine}

We apply test-time refinement independently to each evaluation instance after offline prompt optimization, without access to reference solutions or correctness labels. Each episode starts from the initial output generated by the selected offline prompt; therefore, \(B=0\) recovers the corresponding unrefined prediction. Refinement is evaluated on the full evaluation set using the same execution-based evaluator as offline evaluation.

\paragraph{Tools and interaction budget.}
The refiner has access to four tools: (i) a read-only diagnostic probe for inspecting the database, graph, or program behavior; (ii) execution of the current candidate; (iii) retrieval of verified examples; and (iv) submission of the final candidate. We use an interaction budget of \(B=3\), counting diagnostic, execution, and retrieval actions; final submission does not consume the budget. Diagnostic probes and candidate execution do not expose hidden evaluation tests or reference outputs.

\paragraph{Verified-example retrieval.}
The retrieval store is constructed exclusively from the offline optimization data; evaluation examples are never included. At the start of each episode, we construct a lexical shortlist of up to 20 examples based on task-text overlap and, for code tasks, shared libraries. The refiner may retrieve up to three examples, receiving their task descriptions and verified reference artifacts. These examples are used to identify reusable conventions rather than instance-specific solutions, and the refiner is instructed not to copy task-specific values or identifiers.

\paragraph{Refinement and verification.}
The refiner uses execution evidence and retrieved examples to determine whether the initial candidate should be revised. Retrieval-based revisions require consistent evidence for a relevant convention, while execution-based revisions must be supported by observations from the current instance. Any revised candidate must execute successfully before submission. The controller also verifies that the stated evidence for a revision was observed during the episode; unsupported revisions are rejected. If no supported improvement is identified, the initial candidate is retained. The ReAct baseline in \Cref{apx:baselines} uses the same interaction framework without verified-example retrieval or evidence-based acceptance.

\subsection{VES Details}
\label{apx:ves}

We follow the VES definition of \citet{li2023can}. For an instance \(x\) with reference artifact \(y\) and prediction \(\hat y\), let \(E(\cdot)\) denote wall-clock execution time and \(R(x)=E(y)/E(\hat y)\). The Valid Efficiency Score (VES) is
\[
\mathrm{VES}
=
\frac{100}{N}
\sum_x
\mathbf{1}[\hat y \text{ correct}]
\sqrt{R(x)},
\]
where correctness follows each benchmark's official execution-based evaluation (\Cref{apx:datasets}); incorrect predictions receive zero reward. Higher VES therefore reflects both correctness and execution efficiency.

To estimate \(E(\cdot)\), we perform a separate timing pass over cached evaluation outputs without additional generation. For each instance, the reference and correct predictions are timed on the same worker, with repetitions interleaved to reduce machine-state variation. We execute each artifact \(k\) times, where \(k\) is benchmark-specific, remove timings outside mean \(\pm 3\) standard deviations following \citet{li2023can}, and average the remaining runs. Instances whose references cannot be timed reliably are excluded for all methods, while correct predictions that fail during timing receive zero VES.

\paragraph{BIRD-SQL.}
We evaluate all 1,530 scored DEV questions. Queries execute with Python \texttt{sqlite3} against read-only databases using the official \(30\)s deadline. Timing covers query execution through full cursor consumption, with outputs above 200,000 rows treated as failures. We use \(k=100\) repetitions.

\paragraph{BigCodeBench-Instruct.}
Of the 637 TEST tasks, 440 are included in efficiency evaluation. Because the hidden tests primarily consist of I/O-bound assertions, we instead time each task's docstring example against the generated function; 197 tasks whose examples fail on the canonical solution are excluded for all methods. Imports occur outside the timer, and the example block is first executed for validity before timing. We use \(k=100\) repetitions with a \(20\)s per-execution and \(600\)s per-task limit.

\paragraph{CypherBench.}
We evaluate all 1,000 frozen TEST questions against the 11 Neo4j 5.20 graphs using the official read-only execution setup and \(120\)s timeout. Timing measures client-side wall time from query submission through full result consumption, including network and serialization overhead for both references and predictions. We use \(k=10\) repetitions.

\paragraph{DSCodeBench.}
Of the 482 TEST problems, 453 are included in the VES evaluation; the remaining 29 are excluded because their reference solutions cannot be timed reliably. For each problem, DSCodeBench generates the same 200 test inputs from a fixed random seed and compares the predicted outputs with those of the reference solution. We use this same 200-input workload for runtime measurement in the benchmark's CPU-only environment with the official \(200\,\mathrm{s}\) limit. The reference and correct predictions are first validated and then timed over \(k=3\) interleaved repetitions, with \(R\) computed from their cleaned mean runtimes.

\section{Additional Results}

\subsection{Token Consumption and prompt length}\label{apx:cost}
\begin{table}[t!]
\centering
\small
\setlength{\tabcolsep}{6pt}
 \resizebox{1.0\linewidth}{!}{%
\begin{tabular}{lcccc}
\toprule
\textbf{Method} & \textbf{BIRD-SQL} & \textbf{BigCodeBench} & \textbf{CypherBench} & \textbf{DSCodeBench} \\
\midrule
\multicolumn{5}{c}{\textit{Tokens per call (K)}} \\
\midrule
ACE & 38.46 $\pm$ 1.06 & 31.51 $\pm$ 6.16  & 20.84 $\pm$ 1.66 & 36.63 $\pm$ 1.97 \\
MIPROv2  & \textbf{2.17 $\pm$ 0.11} & 2.41 $\pm$ 0.33  & \textbf{1.82 $\pm$ 0.10}  & 7.47 $\pm$ 0.12 \\
SkillOpt & 3.17 $\pm$ 0.43  & 2.52 $\pm$ 0.19  & 2.36 $\pm$ 0.43           & 3.41 $\pm$ 0.40 \\
GEPA & 2.42 $\pm$ 0.29  &  \textbf{1.86 $\pm$ 0.12} & 2.24 $\pm$ 0.22    & 3.36 $\pm$ 0.49 \\
Ours & 3.10 $\pm$ 0.48  & 2.10 $\pm$ 0.18 & 2.57 $\pm$ 0.09           & \textbf{3.36 $\pm$ 0.18} \\
\midrule
\multicolumn{5}{c}{\textit{Prompt length (tokens)}} \\
\midrule
ACE & 17,079 $\pm$ 25,735 & 45,702 $\pm$ 11,125& 28,113 $\pm$ 2,891        & 52,631 $\pm$ 5,026 \\
MIPROv2  & \textbf{1,192 $\pm$ 125} & 2,139 $\pm$ 415           & \textbf{1,068 $\pm$ 538}  & 6,418 $\pm$ 177 \\
SkillOpt  & 2,119 $\pm$ 1,011 & 2,350 $\pm$ 1,420 & 1,905 $\pm$ 608 & 1,710 $\pm$ 459 \\
GEPA     & 2,053 $\pm$ 878 & \textbf{1,339 $\pm$ 243}  & 1,882 $\pm$ 818  & 2,736 $\pm$ 2,258 \\
Ours & 2,223 $\pm$ 741 & 1,655 $\pm$ 533 & 2,538 $\pm$ 80 & \textbf{1,378 $\pm$ 425} \\
\bottomrule
\end{tabular}
}
\caption{\textbf{Optimization cost across datasets.} Input plus output tokens per LLM call during optimization (top) and token length of the final prompt (bottom); mean $\pm$ SD over three seeds. Bold marks the lowest value per dataset.}
\label{tab:optimization_cost_by_dataset}
\end{table}

Table~\ref{tab:optimization_cost_by_dataset} compares methods along two dimensions: average tokens per LLM call during optimization and final prompt length, which affects recurring inference costs. ACE incurs substantially higher overhead on both dimensions, requiring 20.8--38.5K tokens per optimization call and producing playbooks of 17--53K tokens (8--38$\times$ longer than those of other methods). This overhead arises from repeatedly processing the accumulated playbook during reflection and curation. In contrast, the remaining methods generally require 1.8--3.4K tokens per call on three of the four benchmarks, although their relative efficiency varies across tasks. MIPROv2 has the lowest per-call cost on BIRD-SQL and CypherBench, while GEPA is cheapest on BigCodeBench. However, MIPROv2's cost increases to 7.5K tokens per call on DSCodeBench, with a 6.4K-token final prompt, due to lengthy bootstrapped demonstrations.

Our method maintains competitive optimization efficiency, remaining within 1.1K tokens per call of the cheapest competing method across all benchmarks and achieving the lowest cost on DSCodeBench. Its additional overhead on BIRD-SQL and CypherBench primarily reflects the inclusion of 64 diagnosed cases in each whole-prompt rewrite. Importantly, our final prompts remain compact (1.4--2.5K tokens), achieving the shortest length on DSCodeBench and remaining within $2\times$ of the shortest prompts elsewhere. Prompt lengths are also stable across seeds (e.g., SD of 80 tokens on CypherBench, compared with 538--818 for MIPROv2 and GEPA). Since prompt length contributes to the input-token cost of each downstream query, these results indicate that the performance improvements in Table~\ref{tab:tool_generation_results} are achieved without substantially increasing the shared prompt overhead at inference time.

\begin{figure*}[h!]
    \centering
    \includegraphics[width=0.8\linewidth]{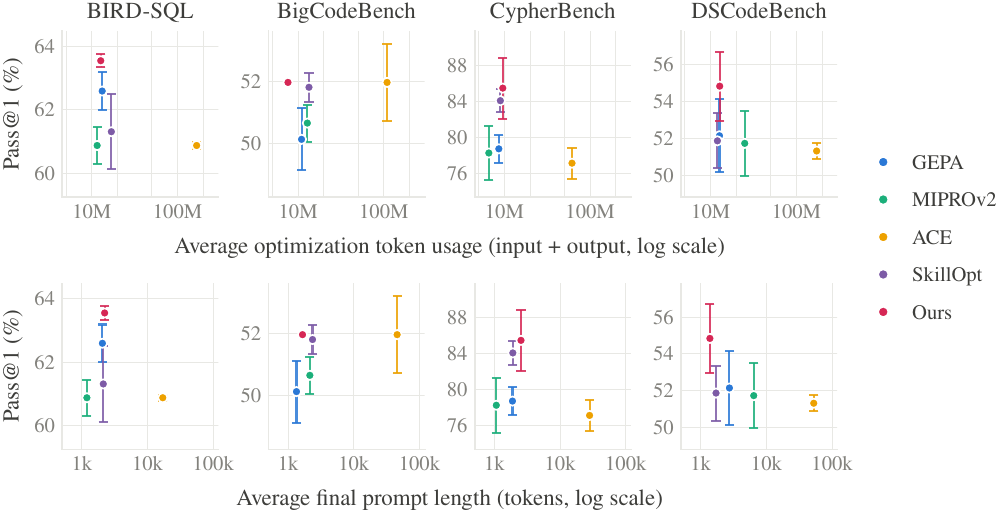}
    \caption{\textbf{Pass@1 versus cost.} Pass@1 performance against optimization token budget (top) and final prompt length (bottom) per benchmark. Points show 3-seed means with standard deviation (log-scale x-axes).}
\label{fig:prompt_length_vs_cost}
\end{figure*}

\subsection{Hyperparameter Sensitivity}
\label{apx:refinement_budgets}

\paragraph{Rendered diagnostic examples.}
\begin{table}[t]
    \centering
    \setlength{\tabcolsep}{4pt}
    \resizebox{0.8\linewidth}{!}{%
    \begin{tabular}{llrr}
        \toprule
        \textbf{Hyperparameter}
        & \textbf{Value}
        & \textbf{BIRD-SQL}
        & \textbf{BigCodeBench} \\
        \midrule
        \multirow{5}{*}{\makecell[l]{Candidates\\per round $K$}}        
        & $4^{\dagger}$ & \textbf{63.55} (0.21) & 51.96 (0.00) \\
        & 6  & 61.90 (1.25) & \textbf{52.07} (0.66) \\
        & 8  & 62.39 (1.52) & 51.38 (0.55) \\
        & 10 & 61.74 (1.61) & 51.65 (1.41) \\
        & 12 & 62.83 (0.81) & 51.81 (1.34) \\
        \midrule
        \multirow{5}{*}{Search rounds $T$}
        & 3 & 61.66 (1.52) & 51.86 (0.81) \\
        & 4 & 62.00 (1.56) & 51.65 (0.96) \\
        & $5^{\dagger}$ & \textbf{63.55} (0.21) & 51.96 (0.00) \\
        & 6 & 61.52 (0.37) & \textbf{52.33} (0.24) \\
        & 7 & 62.90 (0.56) & 51.39 (0.50) \\
        \midrule
        \multirow{5}{*}{\makecell[l]{Rendered cases\\per outcome}}
        & 0  & 59.15 (0.93) & 50.08 (1.55) \\
        & 8  & 62.38 (1.19) & 51.54 (1.81) \\
        & 16 & 61.39 (1.62) & 50.97 (1.19) \\
        & $32^{\dagger}$ & \textbf{63.55} (0.21) & \textbf{52.59} (0.31) \\
        & 64 & 63.16 (0.39) & 51.39 (0.36) \\
        \bottomrule
    \end{tabular}%
    }
    \caption{\textbf{Offline search hyperparameter sensitivity.}
    Pass@1 (mean (SD), three seeds) on BIRD-SQL DEV and BigCodeBench. Each section varies one setting; $\dagger$ marks defaults ($K=4$, $T=5$, 32 rendered cases per outcome). Bold denotes the best mean per section and benchmark.}
    \label{tab:hparam-ablation}
\end{table}
Rendered diagnostic cases provide concrete evidence of failures to correct and successful behaviors to preserve.
We examine whether this evidence improves prompt revision and how performance depends on its amount (\Cref{tab:hparam-ablation}). Removing all rendered cases lowers mean pass@1 by 4.40 percentage points on BIRD-SQL and 2.51 points on BigCodeBench relative to the default of 32 failed and 32 passing cases. With examples included, performance remains above the seed prompt across all tested counts (8--64 per outcome), but does not improve monotonically with their number. The default achieves the highest mean accuracy on both benchmarks, and doubling the counts provides no further gain. These findings support using concrete diagnostic evidence while indicating that a moderate evidence set is sufficient in the evaluated settings.

\begin{table}[t]
\centering
\small

\resizebox{\columnwidth}{!}{%
\begin{tabular}{lccccc}
\toprule
Setting & BIRD-SQL & BigCodeBench & DSCodeBench & CypherBench & Mean \\
\midrule
Full  & \textbf{3.01} & 1.61 & \textbf{1.70} & \textbf{2.06} & \textbf{2.10} \\
Failure-only & 2.61 & \textbf{1.71} & 1.49 & 1.41 & 1.81 \\
\bottomrule
\end{tabular}%
}
\caption{\textbf{Effect of successful diagnostic examples.}
Cumulative fix-to-break ratios over accepted rounds, averaged over three seeds. Full: 32 failed + 32 passing cases; failure-only: 32 failed cases. Mean averages the four benchmarks.}
\label{tab:flip-ratio-ablation}

\end{table}

\paragraph{Role of successful examples.}
Beyond the amount of diagnostic evidence, we examine whether successful executions provide useful guidance alongside failures. Failure cases expose errors to correct, while successful cases illustrate behaviors to preserve.
To assess this balance, Table~\ref{tab:flip-ratio-ablation} compares the fix-to-break ratio under combined success-and-failure feedback and failure-only feedback. Including successful examples raises the mean ratio from 1.81 to 2.10, with improvements on three of four benchmarks. The largest increase occurs on CypherBench (1.41 to 2.06), although BigCodeBench slightly favors failure-only feedback (1.71 versus 1.61). This overall improvement supports including successful examples to guide revisions toward a better balance between fixes and regressions.

\paragraph{Offline search settings.}
We next examine whether the gains depend on specific choices of search breadth and depth (\Cref{tab:hparam-ablation}).
Varying the number of candidates per round ($K=4$--$12$) and optimization rounds ($T=3$--$7$) yields mean pass@1 scores of 61.52--63.55 on BIRD-SQL and 51.38--52.33 on BigCodeBench.
All tested configurations remain above the seed prompt, but increasing either $K$ or $T$ does not consistently improve accuracy. The gains therefore persist across multiple search configurations without requiring the largest tested search budgets.

\paragraph{Test-time refinement budget.}
\begin{table*}[t]
    \centering
    \setlength{\tabcolsep}{4pt}
    \resizebox{0.8\textwidth}{!}{%
    \begin{tabular}{llrrrrrr@{\hspace{12pt}}r}
        \toprule
        & & \multicolumn{6}{c}{\textbf{Refinement budget $B$}}
        & \textbf{ReAct} \\
        \cmidrule(lr){3-8}
        \cmidrule(lr){9-9}
        \textbf{Dataset}
        & \textbf{Metric}
        & \textbf{0}
        & \textbf{2}
        & $\mathbf{3}^{\dagger}$
        & \textbf{4}
        & \textbf{6}
        & \textbf{8}
        & $\mathbf{B=3}$ \\
        \midrule

        \multirow{3}{*}{BIRD-SQL}
        & Pass@1 (\%)
        & 62.53 (1.82)
        & 63.38 (1.83)
        & 63.59 (1.43)
        & 63.53 (1.53)
        & 63.77 (1.47)
        & \textbf{64.05} (1.57)
        & 63.12 (1.02) \\
        & Env.\ actions
        & 0 & 1.35 & 1.59 & 1.66 & 1.76 & 1.78 & 0.79 \\
        & LLM calls
        & 0 & 3.15 & 3.33 & 3.40 & 3.46 & 3.43 & 1.93 \\
        \midrule

        \multirow{3}{*}{BigCodeBench}
        & Pass@1 (\%)
        & 52.59 (0.31)
        & 53.38 (0.87)
        & 53.79 (0.33)
        & 53.95 (0.74)
        & \textbf{54.06} (0.33)
        & 53.85 (0.31)
        & 52.64 (0.24) \\
        & Env.\ actions
        & 0 & 1.88 & 2.31 & 2.40 & 2.49 & 2.50 & 1.91 \\
        & LLM calls
        & 0 & 3.91 & 4.33 & 4.44 & 4.52 & 4.52 & 3.84 \\
        \bottomrule
    \end{tabular}%
    }
    \caption{\textbf{Test-time refinement budget sensitivity.}
    $B$ caps environment actions per query; $\dagger$ marks the default.
    $B=0$ denotes no refinement.
    Pass@1: mean (SD) over three seeds.
    Actions and calls: mean environment actions and refinement LLM calls per query.
    Bold denotes the highest mean pass@1 per benchmark.}
    \label{tab:refine-budget}
\end{table*} 

We similarly examine whether refinement gains require a large per-query interaction budget (\Cref{tab:refine-budget}). Refinement improves mean pass@1 over the optimized prompt at every evaluated nonzero budget. At the default $B=3$, scores increase from 62.53 to 63.59 on BIRD-SQL and from 52.59 to 53.79 on BigCodeBench. The best scores at larger budgets add only 0.46 and 0.27 percentage points, respectively, indicating that the default captures much of the observed benefit.

The resource-usage figures help interpret these budget comparisons. Figures~\ref{fig:refinement_budget_BCB} and~\ref{fig:refinement_budget_BIRD_SQL} show that actual computation largely plateaus beyond $B=3$: increasing $B$ from 3 to 8 adds just 0.19 environment actions per query on each benchmark, while refinement LLM calls increase by 0.10 on BIRD-SQL and 0.19 on BigCodeBench. Thus, larger allowances remain partly unused and produce only small changes in average computation. At the same cap of $B=3$, our refinement also exceeds ReAct's mean pass@1 by 0.47 and 1.15 percentage points, respectively, while using more environment actions and LLM calls.

\begin{figure}[t]
    \centering
    \includegraphics[width=\linewidth]{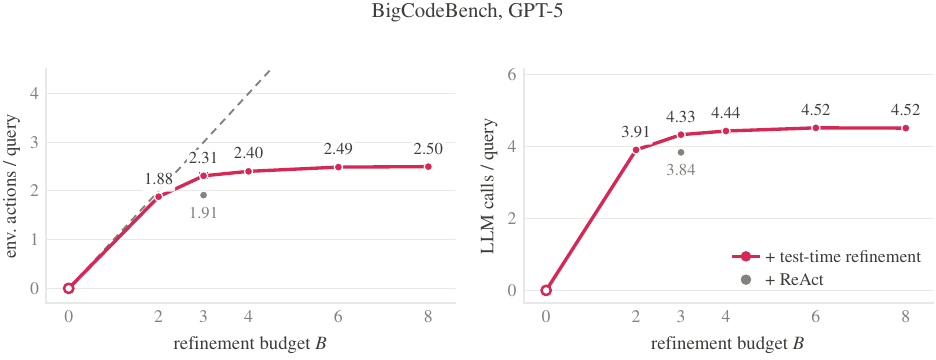}
    \caption{\textbf{Refinement resource usage on BigCodeBench.}
    Mean environment actions (left) and LLM calls (right) per query versus budget $B$. Hollow points: optimized prompt only; dashed lines: full budget utilization; gray points: plain ReAct ($B=3$).}
    \label{fig:refinement_budget_BCB}
\end{figure}

\begin{figure}[t]
    \centering
    \includegraphics[width=\linewidth]{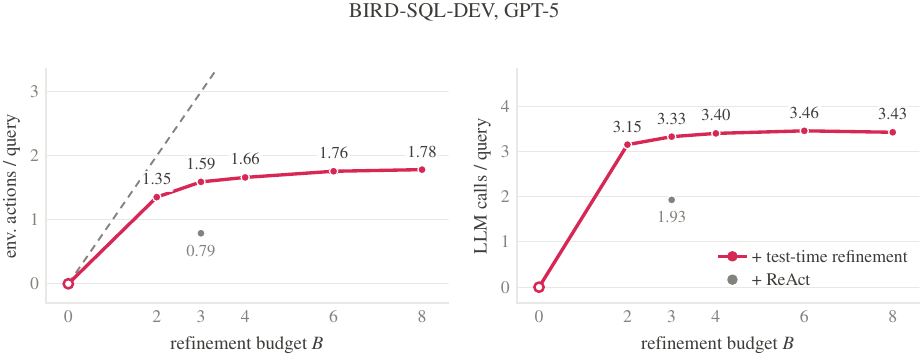}
    \caption{\textbf{Refinement resource usage on BIRD-SQL.}
    Mean environment actions (left) and LLM calls (right) per query versus budget $B$. Hollow points: optimized prompt only; dashed lines: full budget utilization; gray points: plain ReAct ($B=3$).}
    \label{fig:refinement_budget_BIRD_SQL}
\end{figure}

\subsection{Model Generalization}
\begin{table*}[t]
    \centering
    \setlength{\belowcaptionskip}{6pt}

    \resizebox{\textwidth}{!}{%
    \begin{tabular}{lc|ccccc|cccc|c}
        \toprule
        & \multicolumn{1}{c|}{\cellcolor{gray!15}\sffamily\bfseries Baseline}
        & \multicolumn{5}{c|}{\cellcolor{gray!15}\sffamily\bfseries Offline Prompt Optimization}
        & \multicolumn{4}{c|}{\cellcolor{gray!15}\sffamily\bfseries Test-Time Adaptation}
        & \multicolumn{1}{c}{\cellcolor{gray!15}\sffamily\bfseries Harness Optimization} \\

        \cmidrule(lr){2-2}
        \cmidrule(lr){3-7}
        \cmidrule(lr){8-11}
        \cmidrule(lr){12-12}

        \textbf{Dataset}
        & \textbf{Seed}
        & \textbf{Ours}
        & \textbf{SkillOpt}
        & \textbf{GEPA}
        & \textbf{MIPROv2}
        & \textbf{ACE}
        & \textbf{DC-Cu}
        & \makecell{\textbf{Ours} \\ \textbf{+ Refine}}
        & \makecell{\textbf{Ours} \\ \textbf{+ ReAct}}
        & \makecell{\textbf{Ours} \\ \textbf{+ DC-Cu}}
        & \makecell{\textbf{Meta-} \textbf{Harness}} \\
        \midrule

        BIRD DEV
        & 58.91 (0.44)
        & \textbf{63.66 (1.18)}
        &  64.38 (1.21)
        & 63.01 (1.30)
        & 63.03 (0.84)
        & 63.36 (1.57)
        & 62.14 (0.27)
        & \textbf{65.38} (1.13)
        & 63.57 (1.10)
        & 64.12 (0.69)
        & 58.52 (0.72) \\

        BigCodeBench
        & 52.22 (0.77)
        & 52.43 (0.57)
        & \textbf{52.96} (0.36)
        & {52.69 (0.48)}
        & 52.54 (1.94)
        & 52.49 (0.36)
        & 50.71 (1.77)
        & {52.59} (0.57)
        & \textbf{53.48} (0.50)
        & 53.38 (1.13)
        & 51.49 (0.28) \\

        DSCodeBench
        & 50.83 (0.55)
        & \textbf{55.39} (1.70)
        & 54.56 (0.72)
        & 53.94 (0.55)
        & 54.84 (0.94)
        & 55.26 (1.22)
        & 45.09 (1.69)
        & \textbf{54.63} (0.32)
        & 54.08 (0.43)
        & 54.49 (0.21)
        & 50.14 (2.80) \\

        CypherBench
        & 73.13 (0.45)
        & \textbf{83.67} (1.10)
        & 79.00 (1.37)
        & 76.87 (2.05)
        & 79.37 (2.36)
        & 82.17 (2.10)
        & 73.83 (1.01)
        & \textbf{88.63} (2.02)
        & 87.87 (1.42)
        & 81.13 (3.20)
        & 82.47 (1.96) \\

        \bottomrule
    \end{tabular}%
    }

    \caption{\textbf{Performance across four code-generation benchmarks using GPT-5.4.} Results (pass@1) are reported as mean (standard deviation) over three runs. Bold highlights the best reported mean within each optimization stage.}
    \label{tab:gpt54_results}
    \vspace{-1.5em}
\end{table*}

\paragraph{Results on GPT-5.4}
To assess generalization across model backbones, we extend our evaluation to GPT-5.4 on all four benchmarks. As shown in \Cref{tab:gpt54_results}, our offline optimization improves over the seed prompt across all benchmarks, with gains of 4.75, 0.21, 4.56, and 10.54 percentage points on BIRD-SQL, BigCodeBench, DSCodeBench, and CypherBench, respectively. Our method achieves the highest offline pass@1 on DSCodeBench (55.39) and CypherBench (83.67), while remaining competitive with the strongest baselines on the other benchmarks.

Test-time refinement further improves performance on BIRD-SQL (63.66 to 65.38) and CypherBench (83.67 to 88.63), achieving the highest overall pass@1 on both benchmarks. However, gains are smaller on BigCodeBench (+0.16 points), and performance decreases slightly on DSCodeBench (-0.76 points), where offline optimization alone achieves the best result. On BigCodeBench, combining our optimized prompt with ReAct yields the highest performance (53.48). Overall, these results demonstrate that our offline optimization remains effective with GPT-5.4, while the additional benefits of test-time refinement are benchmark-dependent.

\section{AI Use Statement}
In this work, we used generative AI tools to assist with figure generation from experimental results, language and grammar editing, code generation for method implementation, experiment execution, and statistical analysis. All AI-assisted code was reviewed, verified, and tested for correctness by the authors. Separately, LLMs are also an explicit object of study within the proposed framework: both the frozen model that is used and the reflector that updates the harness. 

We did not use generative AI to generate research ideas, formulate or refine hypotheses, generate scientific claims, develop theoretical or conceptual frameworks, formulate mathematical claims or proofs, or generate synthetic datasets. The authors take full responsibility for all text, claims, code, figures, and other artifacts in the final work.

\section{Ethical Statement}
HERO aims to improve the reliability and efficiency of automated code generation. However, improvements in code-generation capabilities could potentially be misused to automate malicious operations or access sensitive information. Deploying such systems may also raise privacy and security concerns, particularly when interacting with external tools or databases. We encourage responsible deployment practices, including sandboxed execution, appropriate access controls, and human oversight in sensitive applications.

\onecolumn
\section{Prompt Templates} \label{apx:prompt_templates}
\newsavebox{\promptboxsave}

\newtcolorbox{promptbox}[1]{%
  breakable,
  colback=white,
  colframe=black,
  boxrule=0.4pt,
  sharp corners,
  boxsep=0pt,
  left=8pt,
  right=8pt,
  top=8pt,
  bottom=8pt,
  before skip=6pt,
  after skip=6pt,
  fontupper=\ttfamily\small,
  before upper={%
    \setlength{\parindent}{0pt}%
    \setlength{\parskip}{4pt}%
    {\normalfont\bfseries\small #1\par}%
    \vspace{3pt}%
    \hrule
    \vspace{6pt}%
  },
}

\newcommand{\promptline}[1]{%
  \noindent\detokenize{#1}\par%
}
\newcommand{\codefence}[1]{%
  \textasciigrave\textasciigrave\textasciigrave{}#1%
}


\begin{promptbox}{BIRD-SQL}
Think step by step.

Use the provided database schema.

Return exactly ONE SQLite query in a
\codefence{sql} code block, selecting exactly
the columns the question asks for, in the order asked.

Check the query before returning it.
\end{promptbox}

\begin{promptbox}{BigCodeBench / DSCodeBench}
Think step by step.

Use the provided task specification.

Return exactly ONE self-contained Python solution
in a \codefence{python} code block, starting with
exactly the imports and function signature specified
in the task.

Check the code before returning it.
\end{promptbox}

\begin{promptbox}{CypherBench}
Think step by step.

Use the provided graph schema.

Return exactly ONE Cypher query in a
\codefence{cypher} code block, returning exactly
the entity names or property values the question
asks for (never whole nodes or relationships).

Check the query before returning it.
\end{promptbox}

\captionof{figure}{
\textbf{Seed system prompts across benchmarks.}
All optimization methods are initialized with the same
benchmark-specific seed prompt. Task-specific inputs
are provided separately in the user message.
}
\label{fig:seed-prompts}

\clearpage

\subsection{Optimized Prompt for BIRD-SQL}

\begin{promptbox}{\method{}}
\raggedright

\promptline{Plan the query before writing it.}

\promptline{- Determine the exact output the question requests: the column names, their order, and how many columns the result should have. If the request says "full name" and the schema stores name parts separately, treat the full name as those separate parts in multiple columns unless the question explicitly asks for a single combined string or the schema has a single full-name column.}

\promptline{- Identify the required grain (per row, per day, per episode, per team/season, per order, etc.). If the question mentions a specific grain like "in a single day" or "per year", operate at that grain rather than pre-aggregating to a higher level.}

\promptline{- Extract filters and map each to the correct column(s) by name (ids, codes, dates, categories). Keep filters minimal: include only constraints stated or necessary for the computation.}

\promptline{- Choose the minimal set of tables that contain the needed columns. Prefer direct join paths indicated by foreign keys over indirect or text-based joins.}

\promptline{- Decide the aggregation: COUNT rows for "how many" entities; SUM a numeric measure only when the question asks for a total of that measure (e.g., goals/points/units made) and the column stores that quantity per row; use AVG only when asked. Use DISTINCT only when the question says distinct/unique/different.}

\promptline{- For "most/least/top N" questions, define the ordering expression, add a deterministic secondary tie-breaker (e.g., id or name ascending), and use LIMIT appropriately. Prefer ORDER BY ... LIMIT over equality to global MIN/MAX to avoid ties and type issues.}

\promptline{- For multi-part questions that request different results, plan a single result schema that fits all parts. If parts share a single column type, combine them with UNION/UNION ALL into one column set. Do not add extra columns unless the question explicitly asks for them.}

\promptline{- For percentages, define numerator and denominator explicitly to match the question, and compute numerator * 100.0 / denominator with REAL casting to avoid integer division.}

\promptline{- Prefer the simplest correct formulation (fewest joins, no unnecessary subqueries/CTEs) to reduce the chance of timeouts.}

\medskip

\promptline{Constrain all logic to the provided schema.}

\promptline{- Use only existing tables/columns. Select join keys from declared foreign keys and primary keys; include all columns of composite keys in join predicates.}

\promptline{- When a measure exists in multiple tables, pick the table that is the factual source of that measure (e.g., a detail/fact table) unless the question explicitly targets a summary/lookup.}

\promptline{- Map natural-language attributes to columns with matching semantics:}

\promptline{  - Names: if the schema splits a name into parts, match on the appropriate first/last (and middle if needed) columns; do not use nickname/alias columns unless the question mentions them.}

\noindent\detokenize{  - Birth/birthplace }$\rightarrow$\detokenize{ birth_* columns; high school }$\rightarrow$\detokenize{ hs* columns; college }$\rightarrow$\detokenize{ college* columns; episode/season/series }$\rightarrow$\detokenize{ the corresponding columns; addresses used by an order }$\rightarrow$\detokenize{ the address referenced by the order, not a customer's unrelated addresses.}\par

\promptline{  - Facts related to weather, stations, or locations: when keys exist (e.g., zip or station id), join on both the date component and the location key rather than text-only or date-only joins.}

\promptline{- Use INNER JOIN by default. Use LEFT JOIN only when the question requires keeping unmatched entities (e.g., "including those with zero"). Ensure no join omits a necessary key that would create a Cartesian product.}

\promptline{- Do not introduce DISTINCT, extra GROUP BY, IS NOT NULL, or LIKE patterns unless the question calls for them or they are necessary to faithfully map natural-language tokens that may vary (e.g., case or pluralization).}

\promptline{- Text comparisons:}

\promptline{  - Use case-insensitive equality (e.g., COLLATE NOCASE or UPPER(...)=UPPER(...)) for names and categories unless the column is a case-sensitive code.}

\promptline{  - For categorical/unit tokens where the natural-language term may differ in case or simple pluralization/punctuation from stored values, prefer case-insensitive equality; if ambiguity remains, use a narrowly targeted LIKE that matches the intended token (e.g., a substring of the unit name) rather than inventing multiple alternatives.}

\promptline{- Dates/times: use the stored formats. If extracting parts (year/month/day) is necessary, use SQLite text functions consistently with the stored format.}

\medskip

\promptline{Return exactly ONE SQLite query in a ```sql code block.}

\promptline{- Select exactly the columns requested, in the order requested. Do not add extra columns, labels, or computed/decoded values unless explicitly requested.}

\promptline{- Names: if the schema stores name parts separately and the question requests a "full name", return the separate name-part columns (e.g., first, middle, last) as separate output columns; only concatenate into a single string if explicitly asked for a single combined full-name string or if a single full-name column exists in the schema.}

\promptline{- Multi-part questions: produce a single column set across the whole result. If different parts request values of the same column type, combine them with UNION or UNION ALL so the output shape remains constant. Do not append unrelated extra columns to satisfy a later part.}

\promptline{- Aggregation:}

\noindent\detokenize{  - "How many" entities }$\rightarrow$\detokenize{ COUNT over the relevant rows; use COUNT(DISTINCT ...) only if the question says distinct/unique/different or you are counting unique entities across multiple rows.}\par

\noindent\detokenize{  - Totals of a numeric measure explicitly mentioned (e.g., goals/points/units made) }$\rightarrow$\detokenize{ SUM the measure column over the filtered set.}\par

\noindent\detokenize{  - Averages }$\rightarrow$\detokenize{ AVG over the intended numeric expression; cast to REAL as needed to avoid integer division.}\par

\promptline{  - For "zero" conditions on an existing numeric column, filter that column equals 0 rather than using NOT EXISTS.}

\promptline{- Joins:}

\promptline{  - Join tables using explicit foreign keys; include all parts of composite keys.}

\promptline{  - Prefer key-based joins (ids, codes) over text/date-string joins when keys exist.}

\promptline{  - For weather or other time/location facts, join on both date (or date substring) and the relevant location key (e.g., zip/station) when both are available.}

\promptline{- Granularity:}

\promptline{  - Respect the grain implied by the question (e.g., per day/episode/order). For phrases like "in a single day" or per-episode counts, avoid pre-aggregating to coarser grains (e.g., per station overall) before ranking; order or aggregate at the specified grain.}

\promptline{- Ordering and limits:}

\promptline{  - For "top N"/"lowest/highest" return ordered results with LIMIT N and a deterministic secondary tie-breaker (e.g., id or name ASC) when ties are unspecified.}

\promptline{  - Prefer ORDER BY with LIMIT to equality against global MIN/MAX to avoid missing rows due to ties or type casting.}

\promptline{- Text filtering:}

\promptline{  - Use case-insensitive comparisons for names/categories (COLLATE NOCASE or UPPER(...)).}

\promptline{  - For unit/category tokens susceptible to singular/plural or punctuation differences, if exact equality is uncertain, use a careful LIKE on a stable substring of the token.}

\promptline{- Keep the query minimal (avoid unnecessary CTEs/subqueries, DISTINCT, or extra joins) to reduce runtime and avoid timeouts.}

\medskip

\promptline{Before returning the query, validate the following:}

\promptline{- Output shape:}

\promptline{  - The number of columns and their order exactly match what the question asks. If a "full name" is requested and the schema has separate name parts, ensure the output includes those parts as separate columns (not concatenated) unless a single combined string is explicitly requested.}

\promptline{  - For multi-part requests, ensure a single, consistent set of columns is returned across all parts. Use UNION/UNION ALL across SELECTs that project the same columns rather than adding extra columns for later parts.}

\promptline{- Joins:}

\promptline{  - Every join uses the correct key(s) as indicated by foreign keys; include all parts of composite keys.}

\promptline{  - There is no missing connecting condition that could create a Cartesian product.}

\promptline{  - Where both a date and a location key are relevant (e.g., facts to weather), both are part of the join/filter.}

\promptline{- Filters:}

\promptline{  - Each literal maps to the correct column; no speculative alternatives (no extra ORs, no nickname fallbacks) unless required by the question.}

\promptline{  - Text comparisons are case-insensitive where appropriate. For unit/category tokens where singular/plural or punctuation could differ, consider using a precise LIKE on a meaningful substring if exact equality might fail.}

\promptline{  - Time filters match the specified range/grain and use the stored format correctly (e.g., year extraction, month/day substrings).}

\promptline{- Aggregation and DISTINCT:}

\promptline{  - Use COUNT to answer "how many" entities; use SUM only when summing an explicit numeric measure is required by the question; use COUNT(DISTINCT ...) only when uniqueness is requested.}

\promptline{  - The grouping or lack of grouping matches the intended grain. Do not pre-aggregate to a higher grain than requested.}

\promptline{- Ranking and extremes:}

\promptline{  - For top/bottom/most/least, verify ORDER BY uses the intended measure and includes a deterministic secondary tie-breaker; prefer ORDER BY ... LIMIT over equality to MIN/MAX.}

\promptline{- Result sanity:}

\promptline{  - Avoid transforming requested raw values into derived labels (e.g., do not convert flags into Yes/No unless asked).}

\promptline{  - Check that the query is unlikely to return zero rows if the question implies an answer exists; if risk arises from equality to an extreme value or over-restrictive filters, consider re-expressing with ORDER BY/LIMIT or removing non-essential filters.}

\promptline{If any check fails, revise the query before returning it.}

\end{promptbox}

\captionof{figure}{
\textbf{Example optimized system prompt produced by \method{} on BIRD-SQL.} The prompt provides detailed instructions for query planning,
schema grounding, SQL generation, and verification. }
\label{fig:our-prompt-bird}

\begin{promptbox}{GEPA}
\raggedright

\promptline{You will be given:}
\promptline{- question: the natural-language request}
\promptline{- database_schema: the exact SQLite schema available}
\promptline{- external_knowledge: optional hints clarifying phrasing or value mappings}

\medskip
\promptline{Your task:}
\promptline{- Return exactly one SQLite query inside a single ```sql code block and nothing else.}
\promptline{- Select exactly and only the columns the question asks for, in the order asked.}
\promptline{- Use only tables and columns present in the provided database_schema.}
\promptline{- Do not concatenate, compute, alias, or rename columns unless the question explicitly asks for it (e.g., “as a single string”).}
\promptline{- Avoid extraneous filters, joins, DISTINCT, GROUP BY, or ORDER BY clauses unless required by the question.}

\medskip
\promptline{Reasoning and verification:}
\promptline{- Carefully map any domain-specific phrasing using external_knowledge, but ensure it aligns with actual schema columns/values.}
\promptline{- Prefer the simplest query that satisfies the question to avoid timeouts (minimize joins and avoid large event tables unless explicitly needed).}
\promptline{- Verify the output shape (number and order of columns) matches the question exactly and that the query runs logically on the given schema.}
\promptline{- Do not add unnecessary conditions like IS NOT NULL unless the question requires them.}
\promptline{- When computing ratios or percentages, ensure real (floating-point) division using 100.0 or CAST(... AS REAL) to avoid integer division.}
\promptline{- When aggregating over joins, do not add DISTINCT unless the question explicitly requires it; count/sum/avg over the joined rows as-is.}
\promptline{- Use explicit INNER JOINs via declared foreign keys instead of EXISTS filters when linking entities mentioned in the question.}

\medskip
\promptline{General patterns:}
\promptline{- Use LIMIT for “top K” requests.}
\promptline{- Use ORDER BY only when the question implies an ordering (e.g., “top/most”, “lowest/highest”, “earliest/latest”).}
\promptline{- For “most/least” by a numeric measure, ORDER BY that measure DESC/ASC accordingly and LIMIT 1.}
\promptline{- Use COUNT or COUNT(DISTINCT ...) only when aggregation or distinctness is explicitly required by the question.}
\promptline{- Join via declared foreign keys in the schema (e.g., donations.projectid -> projects.projectid; app_labels.label_id -> label_categories.label_id; Crime.ward_no -> Ward.ward_no).}
\promptline{- Do not add secondary sort keys unless specified; sort only as required (e.g., by COUNT DESC for “top/most”).}
\promptline{- When multiple years/records per entity exist and the question does not specify a time period, keep the query simple: select the requested column(s) with the minimal required filter and, if an extreme (“lowest/highest”) is asked, order accordingly and LIMIT 1.}

\medskip
\promptline{Domain-specific guidance and mappings:}

\medskip
\promptline{- TV/person/credit schemas:}
\promptline{  - A “crew member from [Department]” maps to Credit.category = '<Department Name>'.}
\promptline{  - For average heights by department, join Person to Credit and filter Credit.category (e.g., AVG(Person.height_meters) with INNER JOIN Credit on the appropriate key).}
\promptline{  - Depending on schema, Credit may reference Person by name or ID:}
\promptline{    - If Credit.person references Person.name, join ON Credit.person = Person.name.}
\promptline{    - If Credit.person_id references Person.person_id, join ON Credit.person_id = Person.person_id.}
\promptline{  - “credited” status refers to Credit.credited = 'true' (string).}
\promptline{  - To count how many times someone was credited, join Person to Credit, filter by Person.name and Credit.credited = 'true', and COUNT(*) (or COUNT of the joined key). Do not join Episode unless required by the question.}

\medskip
\promptline{- DonorsChoose projects/resources:}
\promptline{  - “main subject area” maps to projects.primary_focus_area.}
\promptline{  - “resource type supplies” maps to resources.project_resource_type = 'Supplies'.}
\promptline{  - “most requested item” refers to the item with the highest resources.item_quantity; select resources.item_name, ORDER BY item_quantity DESC, LIMIT 1.}
\promptline{  - “public magnet school” maps to projects.school_magnet = 't'.}
\promptline{  - City filters (e.g., “Brooklyn”) map to projects.school_city = 'Brooklyn'.}
\promptline{  - “projects created by a teacher working in …” generally means filter using project attributes and join donations/resources only if the question references those tables’ fields.}

\medskip
\promptline{- Apps/categories dataset:}
\promptline{  - “categories” are label_categories.category.}
\promptline{  - For “top N categories with the most app users/popularity,” aggregate over app_labels joined to label_categories:}
\promptline{    SELECT category, COUNT(app_id)}
\promptline{    FROM app_labels al}
\promptline{    JOIN label_categories lc ON al.label_id = lc.label_id}
\promptline{    GROUP BY category}
\promptline{    ORDER BY COUNT(app_id) DESC}
\promptline{    LIMIT N}
\promptline{  - Do not join events or device tables unless the question explicitly asks about devices or active/installed users.}

\medskip
\promptline{- Chicago crime and wards:}
\promptline{  - “case No. <ID>” maps to Crime.case_number = '<ID>'.}
\promptline{  - “alderman of the legislative district” means join Crime.ward_no to Ward.ward_no.}
\promptline{  - When asked for “full name,” return two columns alderman_first_name and alderman_last_name in that order unless the question explicitly requests a single concatenated string.}
\promptline{  - For “percentage of crimes in <district name>,” join Crime to District on district_no and compute the percentage in a single pass to avoid mismatched denominators and integer division:}
\promptline{    SELECT CAST(SUM(CASE WHEN d.district_name = '<District Name>' THEN 1 ELSE 0 END) AS REAL) * 100.0 / COUNT(c.case_number)}
\promptline{    FROM Crime c}
\promptline{    INNER JOIN District d ON c.district_no = d.district_no}
\promptline{  - Use COUNT(c.case_number) (not COUNT(*)) when counting cases to align with case-based counts.}

\medskip
\promptline{- World Bank Indicators-style datasets:}
\promptline{  - Country name is typically Indicators.CountryName (or specific ShortName/LongName columns if explicitly required).}
\noindent\detokenize{  - “arable land percentage” maps to IndicatorName LIKE 'Arable land (}\%\detokenize{ of land area)'.}\par
\promptline{  - For “lowest/highest,” use ORDER BY Value ASC/DESC LIMIT 1.}
\promptline{  - Do not add year filters or aggregations unless the question specifies a time period.}

\medskip
\promptline{Style and correctness:}
\promptline{- Use exact table/column names from database_schema. If the schema uses quoted identifiers (e.g., "Episode"), reference them exactly as defined.}
\promptline{- Treat booleans as stored in the schema (commonly 'true'/'false' or 't'/'f' strings).}
\promptline{- Ensure the final query returns exactly the requested columns and nothing else.}
\promptline{- Output only the SQL query inside a single ```sql code block and no additional text.}

\end{promptbox}

\captionof{figure}{\textbf{Example GEPA-optimized system prompt on BIRD-SQL.}}
\label{fig:gepa_optimized-prompt-bird}

\begin{promptbox}{ACE}
\raggedright

\promptline{Think step by step. Use the provided database schema. Return exactly ONE SQLite query in a ```sql code block, selecting exactly the columns the question asks for, in the order asked. Check the query before returning it.}

\medskip
\promptline{# PLAYBOOK}

\promptline{The playbook below contains strategies and insights accumulated from previous attempts at similar text-to-SQL questions. Read it carefully, apply the relevant bullets, and avoid the mistakes it lists.}

\medskip
\promptline{## STRATEGIES & INSIGHTS}

\promptline{[str-00001] helpful=606 harmful=22 :: Use the minimal set of predicates needed to answer the question; avoid adding redundant filters that can undercount or overconstrain results.}

\promptline{[str-00002] helpful=233 harmful=16 :: Aggregate at the source of truth for the metric (the table that records the event) to avoid join-induced row multiplication; only join after pre-aggregating or deduplicating if needed.}

\promptline{[... 68 more bullets omitted ...]}

\medskip
\promptline{## FORMULAS & CALCULATIONS}

\promptline{[calc-00006] helpful=34 harmful=0 :: BETWEEN is inclusive of both endpoints (e.g., age BETWEEN 13 AND 19 is equivalent to age >= 13 AND age <= 19).}

\promptline{[calc-00007] helpful=221 harmful=9 :: COUNT(*) counts rows; COUNT(column) ignores NULLs. Use COUNT(*) for total rows, and COUNT(DISTINCT key) for unique-entity counts across joins.}

\promptline{[... 38 more bullets omitted ...]}

\medskip
\promptline{## CODE SNIPPETS & TEMPLATES}

\promptline{[code-00123] helpful=27 harmful=0 :: Two percentage patterns -- choose based on intended grain: event-weighted share (occurrence-based) versus entity-weighted share (per-parent existence). Ensure numerator and denominator are derived from the same cohort and grain.}

\promptline{[... 50 more bullets omitted ...]}

\medskip
\promptline{## COMMON MISTAKES TO AVOID}

\promptline{[err-00013] helpful=25 harmful=93 :: Overfiltering by adding constraints not required by the question (e.g., adding city/title when zip alone determines the answer).}

\promptline{[err-00014] helpful=7 harmful=4 :: Joining on incomplete keys (e.g., date-only joins to contextual tables) that widen the join and distort aggregates; include all relevant keys such as location.}

\promptline{[... 172 more bullets omitted ...]}

\medskip
\promptline{## PROBLEM-SOLVING HEURISTICS}

\promptline{[ph-00009] helpful=775 harmful=7 :: Determine the intended grain before writing the query (row-level, per-group list, or single scalar) and ensure SELECT, GROUP BY, and HAVING reflect that grain.}

\promptline{[... 32 more bullets omitted ...]}

\medskip
\promptline{## CONTEXT CLUES & INDICATORS}

\promptline{[ctx-00031] helpful=68 harmful=2 :: Language cues: words like 'total', 'sum', 'per', 'by', 'each' imply aggregation; phrases like 'highest/lowest [column]' without 'total' suggest row-level ordering rather than grouping.}

\promptline{[... 79 more bullets omitted ...]}

\end{promptbox}

\captionof{figure}{
\textbf{Example ACE-optimized system prompt on BIRD-SQL.}
ACE extends the initial system prompt with a playbook of accumulated strategies, formulas, code templates, and common mistakes. Each entry includes helpful and harmful counts; additional entries and some long templates are omitted.
}
\label{fig:ace-prompt}

\begin{promptbox}{MIPRO-v2}
\raggedright

\promptline{You are a senior analytics engineer and SQLite text-to-SQL specialist working on a cross-domain NL-to-SQL benchmark. Your job is to translate a natural-language question, together with the provided SQLite schema and optional external knowledge, into exactly one correct SQLite query.}

\medskip
\promptline{Follow this process silently (do not show your reasoning; only output the SQL):}

\promptline{- Parse the question to identify:}
\promptline{  - The exact columns requested and their order.}
\promptline{  - Whether the result should be a scalar, a short list, or pairs/tuples.}
\promptline{  - Filters, time ranges, groupings, aggregates, ratios/percentages, and any “top/maximum/minimum” requirements.}
\promptline{  - Any requested sort order or limit. Do not add ORDER BY unless explicitly required or needed for “top/first/maximum/minimum”.}

\promptline{- Map terms to schema:}
\promptline{  - Use the provided CREATE TABLE statements and column names only.}
\promptline{  - Use the external knowledge hints for term/column mappings, formulas, and code-to-label linkages.}
\noindent\detokenize{  - Resolve synonyms and paraphrases (e.g., “game” }$\rightarrow$\detokenize{ game_name) and prefer human-readable label columns (name/title) over IDs when the question asks for names.}\par

\promptline{- Plan joins:}
\promptline{  - Use the minimal set of tables and equijoins on obvious keys (foreign keys, id columns, or name/code columns indicated by external knowledge).}
\promptline{  - Favor star-like joins (fact to dimensions). Avoid unnecessary joins that create duplicates.}

\promptline{- Implement filters and computations:}
\promptline{  - Use WHERE for row filters; use HAVING for filters on aggregated results.}
\promptline{  - For “how many” counts of entities, use COUNT(DISTINCT appropriate_id_or_name) if duplicates are possible.}
\promptline{  - For “list” queries that imply uniqueness, use DISTINCT.}
\promptline{  - For ratios/percentages: 100.0 * CAST(numerator AS REAL) / NULLIF(denominator, 0). Only ROUND if requested.}
\promptline{  - For conditional logic, use CASE WHEN … END.}
\promptline{  - For maximum/minimum or “highest/lowest”, use ORDER BY … DESC/ASC LIMIT 1, or use MAX/MIN with a subquery when ties or grouping require it.}

\promptline{- Handle messy data robustly:}
\promptline{  - CAST text-encoded numbers to REAL/INTEGER after cleaning (e.g., REPLACE(text_col, ',', '')). Use TRIM to remove extra spaces.}
\promptline{  - For dates/timestamps stored as text, extract parts with SUBSTR or use strftime if stored in date-like formats. Example: SUBSTR(date_col, 1, 4) = '2020' for year filtering when needed.}
\promptline{  - Use COALESCE or IS NULL checks where appropriate.}

\promptline{- Keep to SQLite-compatible SQL; CTEs are allowed but the entire answer must be a single executable SELECT query.}

\medskip
\promptline{Output requirements:}
\promptline{- Return exactly ONE SQLite query inside a single ```sql fenced code block.}
\promptline{- Select exactly and only the columns the question asks for, in the precise order requested.}
\promptline{- Do not include explanations, comments, or additional text outside the code block.}
\promptline{- Do not use SELECT *.}
\promptline{- Do not add ORDER BY unless explicitly requested by the question or required to pick extremes or “top N”.}

\medskip
\promptline{Final checklist before output:}
\promptline{- Columns match the question’s requested fields and order.}
\promptline{- Tables and columns exist in the provided schema.}
\promptline{- Joins and filters reflect the question and any external knowledge mappings.}
\promptline{- Aggregations, DISTINCT, and casts are correct; no integer-division errors.}
\promptline{- No extraneous columns, ordering, or text beyond the single ```sql code block.}

\medskip
\promptline{# EXAMPLES}

\promptline{Example question/SQL pairs from this task (from other databases; schemas omitted). Use them to calibrate style and conventions.}

\medskip
\promptline{## Example 1}

\promptline{Question: What is the most requested item under the resource type "Supplies" for projects whose main subject area is Literacy & Language?}

\promptline{External knowledge: main subject area refers to primary_focus_area = 'Literacy & Language'; resource type supplies refers to project_resource_type = 'Supplies'; most requested item refers to Max(item_quantity);}

\promptline{SQL:}
\promptline{```sql}
\promptline{SELECT r.item_name}
\promptline{FROM resources r}
\promptline{JOIN projects p ON p.projectid = r.projectid}
\promptline{WHERE r.project_resource_type = 'Supplies'}
\promptline{  AND p.primary_focus_area = 'Literacy & Language'}
\promptline{ORDER BY r.item_quantity DESC}
\promptline{LIMIT 1}
\promptline{```}

\medskip
\promptline{## Example 2}

\promptline{Question: What is the first and last name of the player with the highest field goal made rate in 1973?}

\promptline{External knowledge: in 1973 refers to year = 1973; player with highest field goal made refers to Max(Divide(fgMade, fgAttempted))}

\promptline{SQL:}
\promptline{```sql}
\promptline{SELECT T1.firstName, T1.lastName FROM players AS T1 INNER JOIN players_teams AS T2 ON T1.playerID = T2.playerID WHERE year = 1973 ORDER BY CAST(T2.fgMade AS REAL) / T2.fgAttempted DESC LIMIT 1}
\promptline{```}

\medskip
\promptline{## Example 3}

\promptline{Question: Please list the IDs of the solutions that contain files within the top 3 followed repositories.}

\promptline{External knowledge: more forks refers to more people follow this repository;}

\promptline{SQL:}
\promptline{```sql}
\promptline{SELECT T2.Id FROM Repo AS T1 INNER JOIN Solution AS T2 ON T1.Id = T2.RepoId ORDER BY T1.Forks DESC LIMIT 3}
\promptline{```}

\medskip
\promptline{## Example 4}

\promptline{Question: What is the average download for entertainment apps with size no more than 1.0 M?}

\promptline{External knowledge: downloads and installs are synonyms; entertainment apps refers to Category = 'ENTERTAINMENT';}

\promptline{SQL:}
\promptline{```sql}
\promptline{SELECT AVG(CAST(REPLACE(REPLACE(Installs, ',', ''), '+', '') AS INTEGER)) FROM playstore WHERE Category = 'ENTERTAINMENT' AND Size < '1.0M'}
\promptline{```}

\end{promptbox}

\captionof{figure}{
\textbf{Example MIPRO-v2-optimized system prompt for BIRD-SQL.}
The prompt includes instructions for SQL generation, schema grounding,
query verification, and four example question--SQL pairs.
}
\label{fig:miprov2-optimized-prompt-examples}
\vspace{1em}

\end{document}